# Modeling spatio-temporal locality in multi-step forecasting of geo-referenced time series

**Annunziata D'Aversa**[1] · **Gianvito Pio**[1,2] · **Michelangelo Ceci**[1,2,3]

**Abstract**

Forecasting future measurements from geographically distributed sensors is essential across many application domains. However, the spatial distribution of these sensors raises multiple challenges, primarily due to spatial autocorrelation phenomena, that introduce inter-dependencies among nearby locations, that cannot therefore be treated independently by learning algorithms. While some existing approaches can capture such phenomena, they generally model the spatial dimension globally across all locations. On the other hand, the method we propose in this paper, called SPALT, focuses on capturing spatial relationships specifically among time series with similar trends, even if these trends occur at different times, thus modeling the spatio-temporal locality. SPALT leverages linear model trees, which allow us to naturally consider the spatial autocorrelation in a local manner: during the tree-building process, the adopted heuristics aim to group time series exhibiting similar trends into the same node, on which additional features considering the spatial dimension are selectively injected. Additionally, we propose a new pruning strategy, based on Reduced Error Pruning (REP), that also considers the spatio-temporal locality during the tree simplification. Designed for a multi-step setting, SPALT provides forecasts for multiple future time steps across multiple sensors simultaneously. The characteristics exhibited by SPALT can provide significant benefits in different domains, where measurements come from geographically distributed sensors. In this paper, we focus on data produced by sensors located in multiple renewable power plants measuring their energy production at regular, short intervals. Experiments on three real-world datasets demonstrate the effectiveness of SPALT in forecasting the production of energy at different time horizons, and its superior performance in comparison with tree-based models and state-of-the-art neural networks that incorporate both temporal and spatial dimensions.

---

---

Extended author information available on the last page of the article

## 1 Introduction

In recent years, the generation of time series data from sensor networks has grown exponentially across several domains. Sensor networks are groups of distributed sensors that track environmental conditions across multiple locations. Continuous data generation allows real-time analysis of patterns and anomalies, as well as the forecasting of future trends. Forecasting the future measurements of multiple geolocated sensors is particularly relevant in several domains, such as (i) transportation, where forecasting traffic conditions allows for alleviating congestion and optimizing transportation systems, or (ii) weather conditions, which forecasting can be useful for public safety from high-impact meteorological events such as flash floods and thunderstorms. Another relevant domain is that of renewable energy, which is the main focus of this paper. In such a domain, predicting the energy produced by power plants allows planning network interventions and optimizing energy distribution.

Although physical models have long been used in the literature (see weather forecasting), machine learning methods can be employed for building predictive models from historical measurements. Indeed, the adoption of such computational methods is continuously gaining importance in real-world applications, where different kinds of models are adopted, such as autoregressive models (Aasim et al., 2019; Bacher et al., 2009), models learned by machine learning algorithms (Hu et al., 2016; Li et al., 2020; Luo et al., 2021), and hybrid models (Jiang et al., 2017). However, one aspect that physical models exploit more effectively than machine learning methods is the spatial position of geo-located sensors. Actually, the spatial dimension is crucial because it may introduce *spatial autocorrelation* phenomena, which refer to dependencies that may exist among observations at nearby geographical locations. Specifically, spatial autocorrelation is based on the principle that spatial closeness can influence the measurements: the Tobler's first law of geography (Tobler, 1970) states that "everything is related to everything else, but near things are more related than distant things". In other words, spatial proximity among sensors can influence the measurements, due to similar conditions (e.g. environmental, meteorological, and topographical conditions) in which they operate. For instance, solar irradiance sensors in close proximity are influenced by cloud cover and local atmospheric conditions, while sensors clustered in a residential neighborhood may all be characterized by the same consumption patterns. Although this phenomenon is well-recognized in spatial data mining for classification and regression tasks (Stojanova et al., 2011), there are only a few studies in the literature that propose forecasting methods accounting for the spatial dimension when analyzing geo-referenced time series (Dowell & Pinson, 2015; Agoua et al., 2018; Li et al., 2023; Khodayar & Wang, 2019; Altieri et al., 2024).

Another crucial aspect to consider when constructing forecasting models is that real-world time series generated by sensor measurements often show a combination of linear and non-linear trends, even at nearby locations. A pertinent example involves measurements influenced by localized weather conditions: a time series generated by a wind power plant situated on a mountain may show non-linear trends absent in the time series of a wind power plant in a nearby valley (or vice versa), despite their spatial proximity. Therefore, capturing both linear and non-linear trends, along with the exploitation of historical data and spatial information, should in principle lead to more accurate predictions. From a methodological perspective, unlike most current time series forecasting methods (Kim et al., 2015; Karaahmetoglu & Kozat, 2020), our approach captures both linear and non-linear phenomena by considering the spatial dimension. We achieve this by proposing an algorithm that learns a specific type of linear model tree. Notably, instead of modeling spatial information globally across all locations, our method focuses on capturing spatial information among instances with similar trends, modeling what we refer to as *spatio-temporal locality*. An additional important aspect in time series forecasting is the *temporal resolution*. Indeed, in the literature, we can find approaches that analyze data ranging from coarse temporal resolutions (e.g., daily, monthly, or yearly measurements) to fine temporal resolutions (e.g., on an hour-by-hour or minute-by-minute basis), in which case they are also referred to as *nowcasting* approaches (Zhang et al., 2018).

In several situations, a fine temporal resolution is accompanied with the need to make the prediction for multiple time steps ahead simultaneously. Some relevant examples come from the energy field: in the electricity market, hour-by-hour predictions are usually needed for the next day (Ceci et al., 2017); in the context of the optimization of the energy distribution, predicting the production of renewable plants is needed for multiple time steps (usually, up to 18) in the future, at a resolution of few minutes. Therefore, the method proposed in this paper will aim to work in the multi-step learning setting.

The method we propose also comes with a novel model simplification strategy inspired by the Occam's principle, which suggests that simple models should be preferred over complex ones when they achieve a similar generalization error (Domingos, 1999). For tree-based models, pruning strategies or early stopping criteria are usually adopted to simplify the model in terms of the number of nodes (Quinlan, 1987). In this work, we propose a novel pruning strategy that leverages the concept of spatio-temporal locality, based on the Reduced Error Pruning (REP) (Quinlan, 1987).

The focus of our experimental evaluation will be the prediction of the energy produced by renewable power plants. Specifically, we aim at learning nowcasting models capable of predicting the energy production for 6, 12 or 18 future time steps, at a fine temporal granularity (from 5 to 15 min, according to the available data). This application domain is motivated by the fact that models able to forecast the energy production play a fundamental role. In long-term scenarios, they can support planning interventions on the network, aiming not only to decrease production costs but also to contribute to the reduction of greenhouse gas emissions. In short-term scenarios, the forecasting of energy production can be useful for performing real-time load balancing actions, including powering on backup plants or drawing energy from customers' accumulators, as well as determining the final electricity clearing price for the energy market in advance.

In summary, the main contributions of this paper are:

- a novel strategy that captures and models the spatial relationships between instances exhibiting similar trends, capturing the spatio-temporal locality within a model able to consider both linear and non-linear dependencies;
- an innovative pruning strategy that leverages the concept of spatio-temporal locality, based on the Reduced Error Pruning (REP), to avoid potential model overfitting as well as to reduce model complexity;
- a thorough time complexity analysis, also demonstrating that the consideration of spatio-temporal locality does not alter the time complexity of the underlying learning algorithms;
- a thorough experimental evaluation on three real-world datasets, demonstrating the ef-

fectiveness of the approach in forecasting energy production of geographically distributed renewable power plants;
- a comprehensive ablation study that assesses the contribution of modeling spatio-temporal locality within linear model trees, and of the proposed pruning strategy, in terms of forecasting errors and model complexity.

The remainder of the paper is organized as follows: Sect. 2 briefly describes the related work. In Sect. 3, we describe in detail the proposed method. In Sect. 4, we describe our experimental evaluation and the obtained results. Finally, Sect. 5 concludes the paper and outlines future work.

## 2 Related work

In the following subsections, we briefly review existing approaches relevant to this paper. First, we examine methods able to build spatio-temporal forecasting models for multiple time series. Then, we focus on forecasting and nowcasting methods designed for or applied to the energy sector.

### 2.1 Spatio-temporal forecasting of multiple time series

In the literature, several forecasting approaches able to model the temporal and the spatial dimensions have been proposed, and applied to several different domains. Methodologically, some approaches proposed in the literature exploit hybrid architectures based on CNN and LSTM (Shi et al., 2015; Yan et al., 2021; Yao et al., 2018; Du et al., 2021). For instance, in Shi et al. (2015) the authors propose a method for 6-minute ahead precipitation nowcasting. Using weather radar data, the authors apply a Convolutional LSTM network, that uses convolutional operations to capture spatial patterns and LSTM units to model temporal dynamics.

In Yao et al. (2018) the authors propose a Deep Multi-View Spatial-Temporal Network (DMVST-Net) for taxi demand prediction. DMVST-Net combines data from three views: temporal, spatial, and semantic. CNN and LSTM are used to learn spatial and temporal dependencies, respectively. On the other hand, the semantic view is captured by *i)* representing different locations as nodes of a graph, *ii)* computing the weights of the edges as the similarity (according to the DTW measure) among time series of the locations, and *iii)* applying a network embedding technique. The latent features of the three views are then fused together through a fully-connected layer. Recently, several approaches based on Graph Convolutional Networks (GCNs) have been proposed, often combined with Recurrent Neural Networks, to simultaneously consider both temporal and spatial dimensions. A relevant example is Graph WaveNet (Wu et al., 2018), a spatio-temporal graph convolutional network for multi-step forecasting, tailored for the prediction of traffic conditions at different locations. It exploits dilated convolution layers to capture temporal dependencies and a self-adaptive adjacency matrix to capture spatial correlations. In Guo et al. (2019), the authors introduce two attention layers into GCNs to capture the dynamic spatio-temporal characteristics of traffic data. They model recent, daily-periodic, and weekly-periodic dependencies by segmenting the time series into three distinct parts. Each segment is processed through a network architecture, consisting of multiple spatio-temporal blocks (spatio-temporal attention and convolution module) and a fully connected layer. The outputs of these networks are then combined to produce the final prediction. It is noteworthy to note that, while the mentioned methods can leverage spatial information, they are unable to model spatio-temporal locality. An initial attempt can be found in Shao et al. (2022), where the authors proposed the method $D^2$STGNN applied to traffic speed forecasting. $D^2$STGNN identifies diffusion signals, which depict how traffic conditions spread across the network, and inherent patterns, such as recurring traffic patterns or daily/seasonal variations. $D^2$ST-GNN adopts a spatio-temporal convolution to capture hidden diffusion time series, and a combination of GRU (for short-term dependencies) and multi-head self-attention mechanism (for long-term dependencies) to model hidden inherent time series. Another method aiming to model the spatio-temporal locality is STAEformer (Liu et al., 2023), that introduces a novel component called *spatio-temporal adaptive embedding* that leads to a significant increase in predictive performances of vanilla transformers. Such a component aims to capture spatio-temporal relationships in an adaptive way, rather than on the basis of a given (static or dynamic) distance matrix. Finally, it is worth mentioning the method HSTGNN (Wang et al., 2025), that is based on a hybrid spatial–temporal graph neural network. For the spatial dimension, HSTGNN leverages time-varying spatial structures and employs a hybrid graph learning approach to learn compound spatial correlations, from both macro and micro perspectives.

Being $D^2$STGNN, STAEformer and HSTGNN conceptually very close to the method proposed in this paper, in our experimental evaluation (see Sect. 4), they will be considered as state-of-the-art competitor approaches.

### 2.2 Forecasting methods in the energy sector

Numerous studies addressing forecasting and nowcasting tasks within the energy sector have been proposed in the literature (Dowell & Pinson, 2015; Li et al., 2023; Khodayar & Wang, 2019; Ceci et al., 2017; D'Aversa et al., 2022; Chen et al., 2019; Tan et al., 2020; Messner & Pinson, 2019). However, only a few studies explore the potential benefits of incorporating the spatial information. In the following, we will focus on methods that specifically take the spatial dimension into account.

In Ceci et al. (2017), the contribution of the temporal and spatial dimensions is considered to predict the hourly energy production of photovoltaic power plants, 24 h ahead. On the other hand, in D'Aversa et al. (2022), the authors propose several learning settings for the prediction of the monthly energy consumption of customers, one year ahead. These works consider a multi-step setting, where the 24 hly predictions (in Ceci et al. (2017)) and the 12 monthly predictions (in D'Aversa et al. (2022)) are returned simultaneously by the model, possibly exploiting dependencies among them. The spatial dimension is considered by resorting to two well-known techniques in spatial statistics: the Local Indicator of Spatial Association (LISA), that represents a local measure of spatial autocorrelation (Anselin, 1995), and the Principal Coordinates of Neighbour Matrices (PCNM), that represent the spatial structure in the data (Dray et al., 2006). Such indicators are used to augment the feature space.

In Dowell and Pinson (2015) the authors propose a method for 5-minute ahead wind power forecasting. The authors capture spatio-temporal dependencies using a method based on a sparse parametrization of VAR models, which selects coefficients that link locations with a spatial co-dependency, discarding those exhibiting weak dependencies. However, VAR models cannot capture complex non-linear dependencies among different variables. In Khodayar and Wang (2019), the authors address the task of 5-minute ahead wind speed forecasting (that can undirectly be exploited to predict wind energy production) using a different approach. They modeled data as an undirected graph where nodes are the wind sites and the edges represent similarities in terms of mutual information computed on the time series, on which a threshold based on the distance is applied. For each site, an LSTM maps the historical data into the temporal feature space. The extracted temporal features are then fed to several spectral-based graph convolutional layers.

In Li et al. (2023) the authors propose a spatio-temporal GCN for the short-term prediction of the energy produced by wind power plants. The authors consider a multi-step setting, where 16 future values (at a 15-minutes interval) are predicted simultaneously. Another work that exploits GCNs is Wu et al. (2020), where the authors propose the system MTGNN and apply it to multiple application domains, including energy and traffic speed forecasting. MTGNN employs multiple temporal convolutional networks (TCNs) with various kernel sizes, for learning temporal dependencies at different scales, and a self-adaptive adjacency matrix to capture spatial correlations.

Another relevant example is GAP-LSTM (Altieri et al., 2024), a recent method that tackles the forecasting tasks in the domains of renewable energy, air pollution, and traffic. The authors propose a novel GCN-LSTM cell that performs graph convolutions at each time step, aiming to consider the spatial information during the whole time series modeling process, in order to extract and preserve complex spatio-temporal patterns that were latent in the original data.

We remark that, although all the mentioned methods are able to model the spatial dimension, none of them is able to model the spatio-temporal locality, but only consider spatial information globally across all locations.

## 3 The proposed method

In this section, we present our novel method called SPALT (Spatiotemporal locality-Aware Pruned Linear model Trees). As introduced in Sect. 1, we aim at *i)* capturing both linear and non-linear dependencies within the data, *ii)* modeling spatio-temporal locality phenomena, and *iii)* possibly reduce the complexity of the learned model, still considering spatio-temporal locality phenomena.

### 3.1 Background

The method SPALT proposed in this paper works on geo-referenced time series data. A time series is a sequence of measurements at different time, typically collected at regular intervals. When the data at hand come from multiple locations, it is important to keep track of, and possibly leverage, the spatial information associated to each time series. The most relevant task that can be solved through time series analysis is the forecasting of future value(s). More formally, given as input $w$ historical values $y_{t-w}, y_{t-w+1}, ..., y_{t-1}$ of the target variable $y$, the task is to predict the value of $y$ for $h$ future time steps, for multiple time series (each associated with its geographic location). When $h = 1$, the task is called *single-step forecasting*, while when $h > 1$, it is usually called *multi-step forecasting* (see Fig. 1 for a graphical example with $w = 12$ historical observations and $h = 6$).

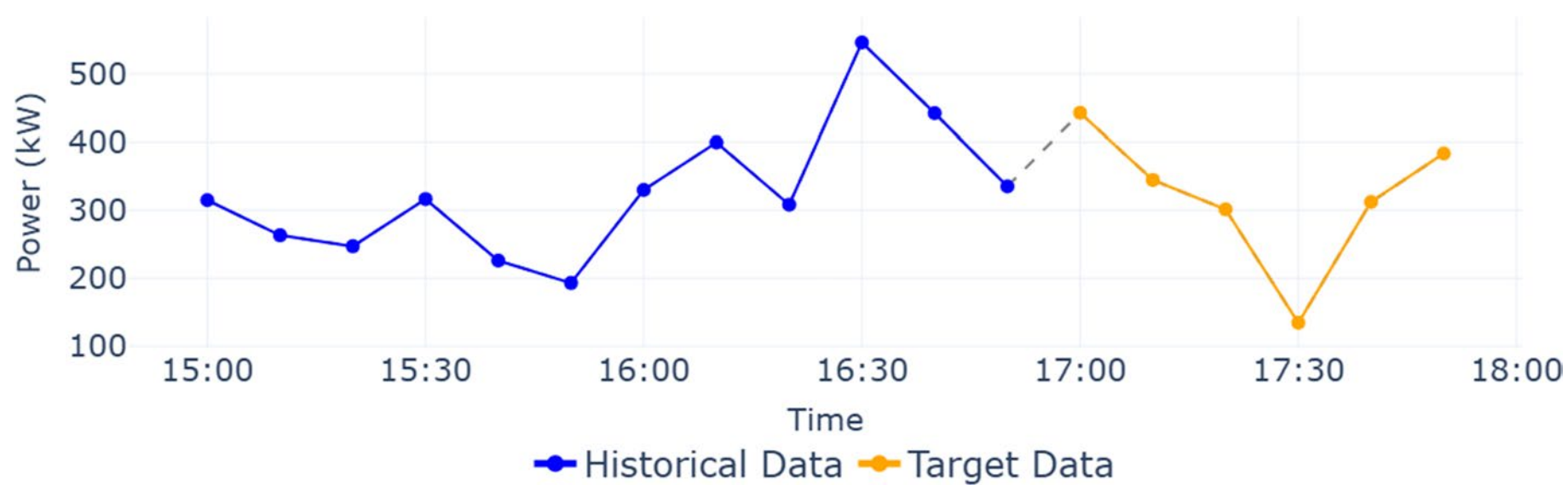

**Fig. 1** An example of a time series with $w = 12$ historical time-steps (in blue) and $h = 6$ time-steps to predict (in orange), related to the energy production from wind power plants measured every 10 min (SDWPF dataset used in our experiments)

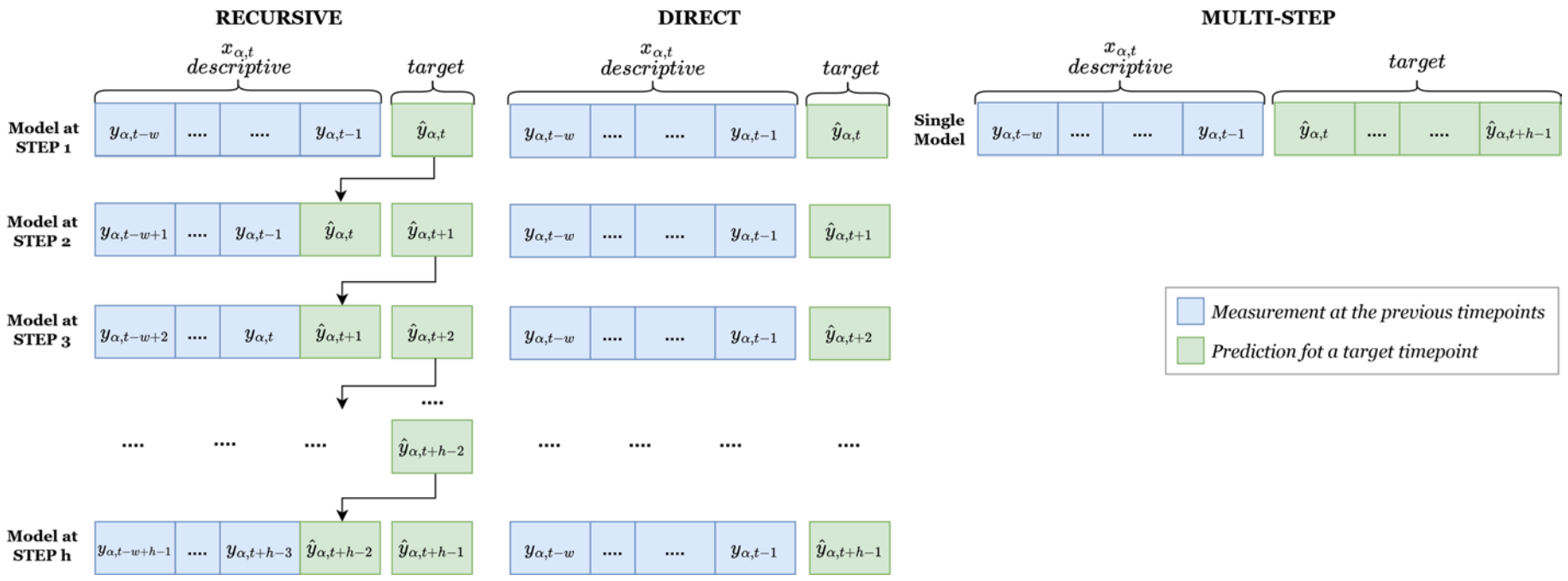

**Fig. 2** A graphical representation of different approaches for multi-step forecasting

According to Taieb et al. (2012), several approaches can be adopted for multi-step forecasting, that can mainly be categorized as *recursive*, *direct*, and *Multi-Input Multi-Output (MIMO)* (see Fig. 2 for a graphical representation of the different approaches). The *recursive* strategy is based on self learning, that is, it re-iterates a single-step ahead predictive model to obtain the desired forecasts: after predicting the next value of the time series, it is fed back as a descriptive variable for the subsequent prediction. The *direct* strategy is based on learning a set of independent predictive models, where the *i*-th model is able to return a prediction for the *i*-th time points in the future. Both *recursive* and *direct* strategies actually appear as single-step approaches applied multiple times to obtain a multi-step prediction. On the other hand, approaches based on the *MIMO* strategy aim to learn one global model that returns a vector of predictions, also possibly taking into account dependencies among future values, that in principle may be beneficial in terms of forecasting accuracy (Bontempi & Ben Taieb, 2011). Given these potential advantages, the method SPALT proposed in this paper falls in the MIMO category.

### 3.2 SPALT

To capture both linear and non-linear dependencies, it is possible to adopt different solutions. Among them, it is worth mentioning methods based on Locally Weighted Projection Regression (LWPR) (Vijayakumar & Schaal, 2000), that combine multiple locally linear models to approximate complex functions, and linear model trees (Quinlan et al., 1992). In this paper, we rely on linear model trees, that combine the ability to model non-linear dependencies of regression trees with that of linear models of properly capturing linear dependencies. Existing methods for the construction of model trees employ a top-down learning process that recursively partitions the training set, analogously to that adopted by conventional tree-based learning algorithms. The main difference between classical regression trees and linear model trees consists in learning a linear model for each leaf node, instead of using a constant value (typically, the mean of the target variable of training instances in a leaf node - see Fig. 3 for an example showing the difference).

Our approach falls in the MIMO category (Taieb et al., 2012), whose goal is to learn a global predictive model that returns the whole vector of predictions, also taking into account the possible dependencies between future values.

Therefore, in our linear model trees, the prediction returned by each leaf node $l$ is based on a multivariate linear function $f_l : \mathbb{R}^w \rightarrow \mathbb{R}^h$ learned from the training instances that reach that leaf node. $f_l(\cdot)$ returns a sequence of $h$ future values for $y$, namely $\langle \hat{y}_t, \hat{y}_{t+1} ..., \hat{y}_{t+h-1} \rangle = f_l(y_{t-w}, y_{t-w+1}, ..., y_{t-1})$. In order to properly learn such linear models, a minimum number of training instances need to fall into each leaf node. Such number of training instances is usually governed by an input parameter $msp \in [0; 1]$, that defines the minimum portion of training instances that should fall into each leaf node. Within each leaf node, the linear model is learned using Ordinary Least Squares which determines the optimal coefficients that minimize the squared differences between predicted and actual values of training instances.

In general, building tree-based models involves recursively identifying the optimal data partitioning using heuristics that evaluate the quality of candidate splits. A split is a test on an input feature, which in our case is of the form $y_{t-j} \leq \theta$, where $j = 1, 2, \ldots, w$ and $\theta$ is a learned threshold. This test partitions the instances of a node into two child nodes based on the test result (true/not_true). In principle, every pair of attribute/threshold should be evaluated to determine the optimal split. To save computational time, in SPALT we adopt the *quantile binning*: for each input feature, instead of evaluating every possible threshold, we identify $b$ quantiles of the distribution of values, that are used as candidate thresholds. This choice drastically reduces the number of thresholds to evaluate for each attribute from (potentially) the number of instances to a constant value $b$ (see Sect. 3.5 for details). A graphical representation of quantile binning is shown in Fig. 4, where data are split into 4 bins, and the quantile boundaries determine the candidate threshold values used for the identification of the optimal split.

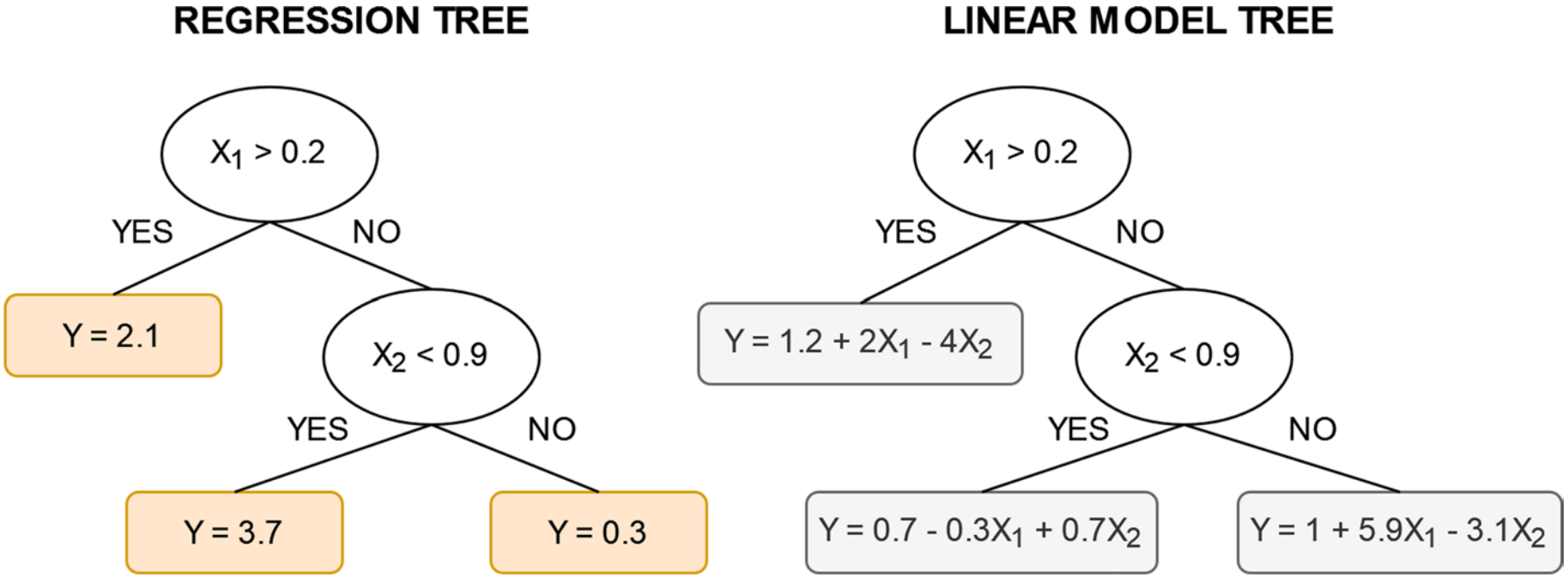

**Fig. 3** An example showing the difference between a classical regression tree (on the left) and a linear model tree (on the right)

As for the adopted heuristic, it must be able to measure how well the split separates data with respect to the target variable. For example, in CART (Breiman et al., 2017), the quality of a split is evaluated by the Mean Squared Error (MSE). In the case of SPALT, we still use MSE, but such a heuristic, used to identify the best split during the tree construction, also needs to consider all the $h$ future time steps. Therefore, it is defined as follows:

$$MSE_l = \frac{1}{h} \cdot \sum_{i=0}^{h-1} \frac{1}{n_l} \cdot \sum_{y_{\beta,t} \in l} (y_{\beta,t+i} - \widehat{y}_{\beta,t+i})^2 \tag{1}$$

where $MSE_l$ is the MSE at node $l$, $n_l$ is the number of instances falling into the node $l$, $y_{\beta,t}$ is the value observed for $y$ at location $\beta$ and time $t$ for a generic instance fallen into the node $l$, and $\hat{y}_{\beta,t}$ is the value predicted for such an instance obtained by fitting a linear model as introduced previously. When a node is split, the MSE is computed for each resulting child node, and the purpose is to minimize the weighted sum (according to the number of instances) of these MSE values. More formally, given $l(y_{t-j}, \theta, left)$ and $l(y_{t-j}, \theta, right)$ the left and right child nodes of the node $l$, respectively, obtained after splitting the instances in $l$ considering the input feature $y_{t-j}$ and the threshold $\theta$, the following heuristics is computed:

$$\frac{1}{n_l} \cdot \left( n_{l(y_{t-j},\theta,left)} \cdot MSE_{l(y_{t-j},\theta,left)} + n_{l(y_{t-j},\theta,right)} \cdot MSE_{l(y_{t-j},\theta,right)} \right),$$

and the optimal pair of $y_{t-j}$ and $\theta$ is selected to minimize such an heuristics.

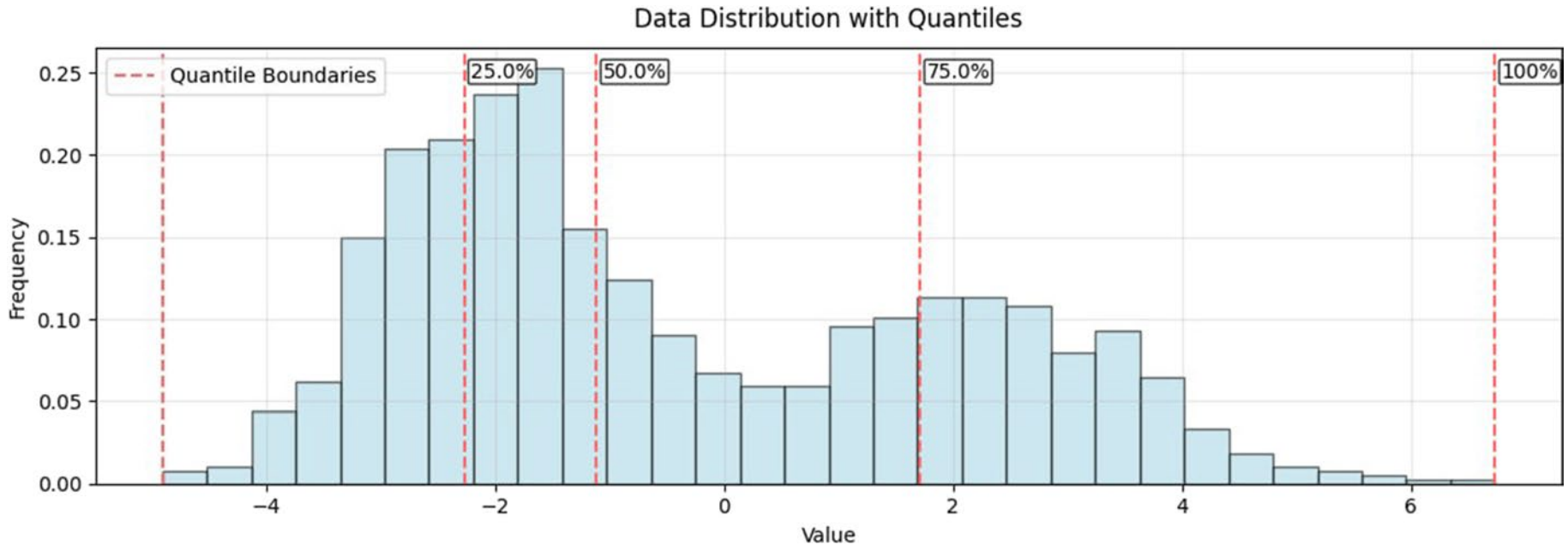

**Fig. 4** A graphical representation of the quantile binning, with $b = 4$. Red-dashed lines represent boundaries at $0\%, 25\%, 50\%, 75\%$, and $100\%$, defining the thresholds for candidate splits

### 3.3 Capturing spatio-temporal locality in SPALT

The literature presents several methods for learning linear model trees (Quinlan et al., 1992; Malerba et al., 2004). As stated in Section 1, this work focuses on learning trees that effectively capture and model spatio-temporal locality. Specifically, we consider the spatial dimension during a post-processing step, aiming to capture spatial relationships among locations within each subset of instances implicitly defined by a leaf node of the model tree.

Intuitively, time series data generated by multiple sensors at different locations are preprocessed using a sliding window approach with a size of $w$. This step aims to generate multiple $w$-dimensional training instances for each sensor. After training a linear model tree, as previously described, we generate a vector of additional features $\mathbf{S}_{\boldsymbol{\alpha},\mathbf{t},\mathbf{l}}$ for each instance that falls into a leaf node $l$, corresponding to the time step $t$ and geographic location $\alpha$. These features are computed as the weighted average of the $w$-dimensional historical observations at the same time step $t$ from other locations in $l$, with weights determined by the spatial proximity between $\alpha$ and the other locations (see Fig. 5). If a given leaf node $l$ contains training instances from only one location, this step is skipped, since no additional features can be injected from other spatio-temporally close instances.

More formally, $\mathbf{S}_{\boldsymbol{\alpha},\mathbf{t},\mathbf{l}}$ is defined as follows:

$$\mathbf{S}_{\boldsymbol{\alpha},\mathbf{t},\mathbf{l}} = \frac{1}{\sum_{\beta \in P_l, \beta \neq \alpha} C[\alpha, \beta]} \cdot \sum_{\beta \in P_l, \beta \neq \alpha} C[\alpha, \beta] \cdot \mathbf{x}_{\boldsymbol{\beta},\mathbf{t}} \tag{2}$$

where:

- $P_l$ is the set of distinct locations of the training instances fallen in the leaf node $l$;
- $\mathbf{x}_{\boldsymbol{\beta},\mathbf{t}}$ is the vector of $w$ historical observations of the location $\beta$ at the time step $t$;
- $C[\alpha, \beta]$ is the spatial closeness between the locations $\alpha$ and $\beta$ computed as follows:

$$C[\alpha, \beta] = 1 - \frac{D[\alpha, \beta]}{max(D)} \tag{3}$$

being $D$ the (geodesic) distance matrix among all the locations.

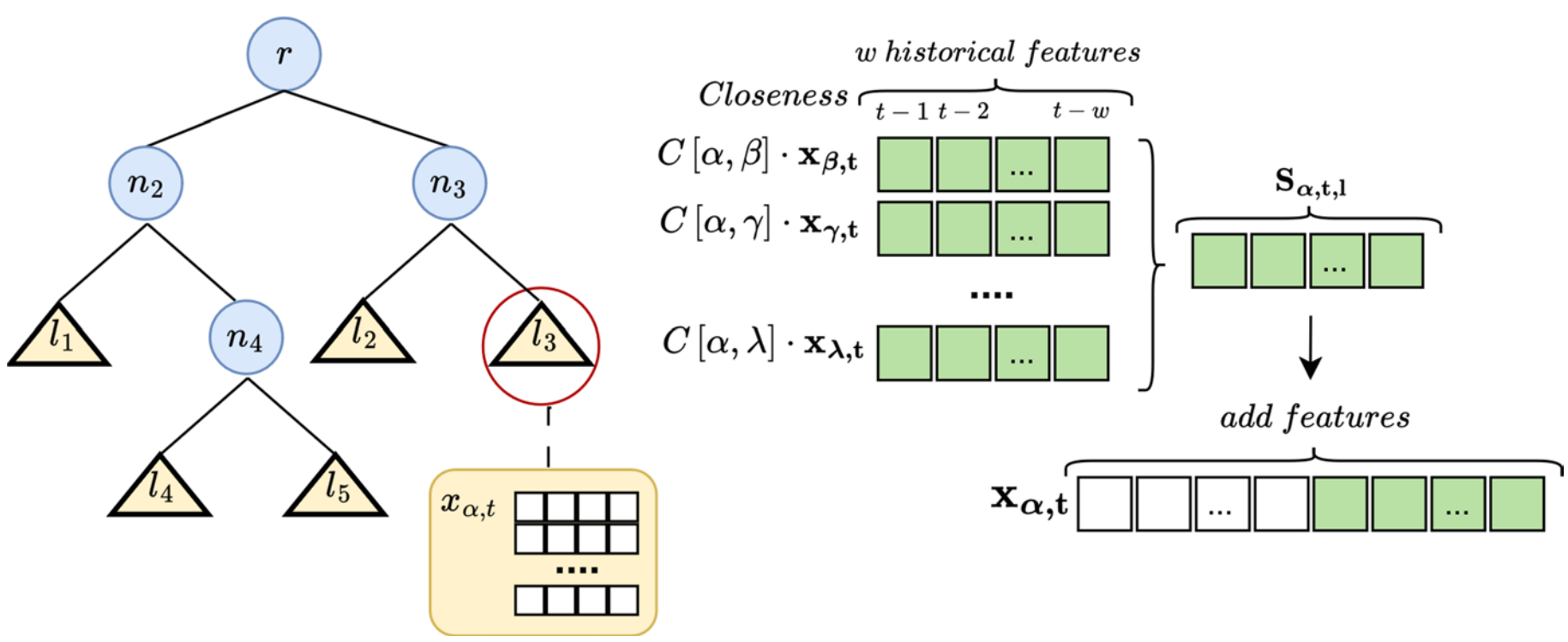

**Fig. 5** An example of computation of the spatio-temporal locality features $\mathbf{S}_{\boldsymbol{\alpha},\mathbf{t},\mathbf{l_3}}$ for the instance $\mathbf{x}_{\boldsymbol{\alpha},\mathbf{t}}$ fallen in the leaf node $l_3$, given the presence of other instances in $l_3$ belonging to the locations $\beta$, $\gamma$, and $\lambda$

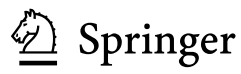

Tree-based models typically use splitting heuristics designed to group training instances with similar historical trends within the same node. By focusing on instances that fall within the same leaf node $l$ and at a specific time step $t$, and weighting their contributions based on spatial proximity, we can enrich the feature set for the linear model originally learned for $l$. This approach allows us to capture both temporal and spatial patterns, enabling the model to account for spatio-temporal locality.

Once SPALT generates and incorporates these additional features for the training instances in a given leaf node, a new linear model is trained. The effectiveness of the added features is then evaluated using a validation set. This process involves comparing two separate linear models, as illustrated in Fig. 6. The first is trained solely on the original features used during the tree construction. The second uses both the original features and the new spatio-temporal features. The model with the lowest validation error is retained for each leaf node. This selective approach ensures the modeling remains aligned with the specific characteristics of the data in each node while avoiding unnecessary complexity. Moreover, incorporating spatio-temporal locality only when it improves validation performance, the method minimizes the risk of overfitting.

As a result, the final tree may include a mix of models: some nodes will use spatio-temporal features, while others will rely exclusively on the original features, depending on the observed advantage of the additional features in each case.

### 3.4 Reduced error pruning based on spatio-temporal locality

As mentioned in Sect. 1, we additionally propose a novel strategy to simplify the obtained linear model trees and, thus, avoid model overfitting. Specifically, we propose a variant of the Reduced Error Pruning (REP) (Quinlan, 1987), that also takes into account the possible contribution of spatio-temporal locality features. The adoption of a pruning strategy also has the indirect positive effect of possibly capturing spatio-temporal correlations at a higher granularity, when it appears to be beneficial for the predictive performance on the validation set.

Methodologically, the original REP strategy begins at the bottom of the tree and moves upward. For each internal node, it compares the error of the unpruned tree with that of a simulated tree, where the subtree rooted at that node is pruned. The subtree is pruned only if the pruned tree performs at least as well as the unpruned tree on the validation set. Our solution also considers the possible contribution provided by the spatio-temporal locality:

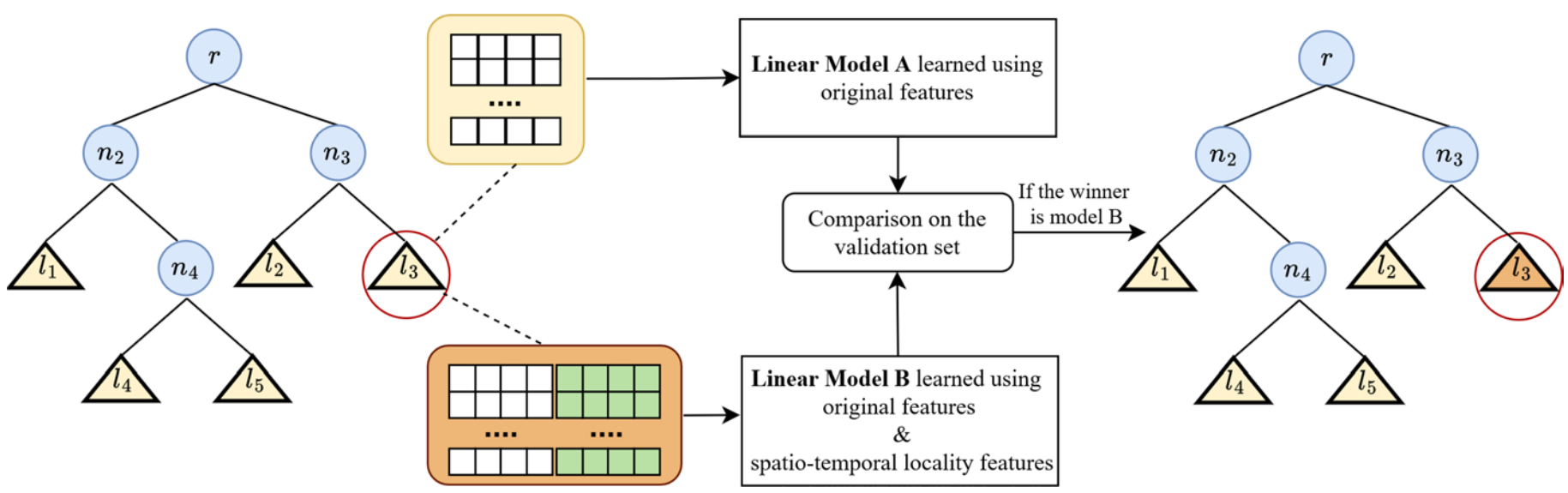

**Fig. 6** Comparison of the predictive performance of two linear models on the training instances of a leaf node: the first model is learned from the original features, while the second is learned from the original features expanded with the features computed according to the spatio-temporal locality

we not only compare the unpruned tree with the pruned tree, but also with the pruned tree where the linear model in the new leaf node (obtained after pruning the subtree rooted on it) also considers spatio-temporal locality features.

In Fig. 7 we report an example. Given the internal node $n_4$, we compare the errors made on the validation set by 3 models:

(i) the model represented by its two children nodes $l_4$ and $l_5$ (see Fig. 7 - left part);
(ii) the model obtained by pruning the subtree rooted in $n_4$ and by learning a new linear model from the training instances falling into it (see Fig. 7 - middle part);
(iii) the model obtained by pruning the subtree rooted in $n_4$ and by learning a new linear model from the training instances falling into it, also considering the spatio-temporal locality features (see Fig. 7 - right part).

If the model *(ii)* or the model *(iii)* leads to an improvement on the validation set, the tree is pruned accordingly. This process continues in a bottom-up fashion until no improvement on the validation set is observed.

A final detail pertains to the validation set, which is the same as the one referenced in Sect. 3.3, but used for a different purpose, namely pruning. This approach eliminates the need to use two separate validation sets.

### 3.5 Time complexity analysis

In this section, we estimate the computational time complexity of SPALT. Let:

- $n$ be the number of training instances;
- $m$ be the number of descriptive variables, that include (or may correspond to) $w$ historical values;
- $b$ be the maximum number of bins for the quantile binning used during the identification of the splits;
- $msp \in ]0;1]$ be the relative minimum number of samples that should fall into each leaf node.

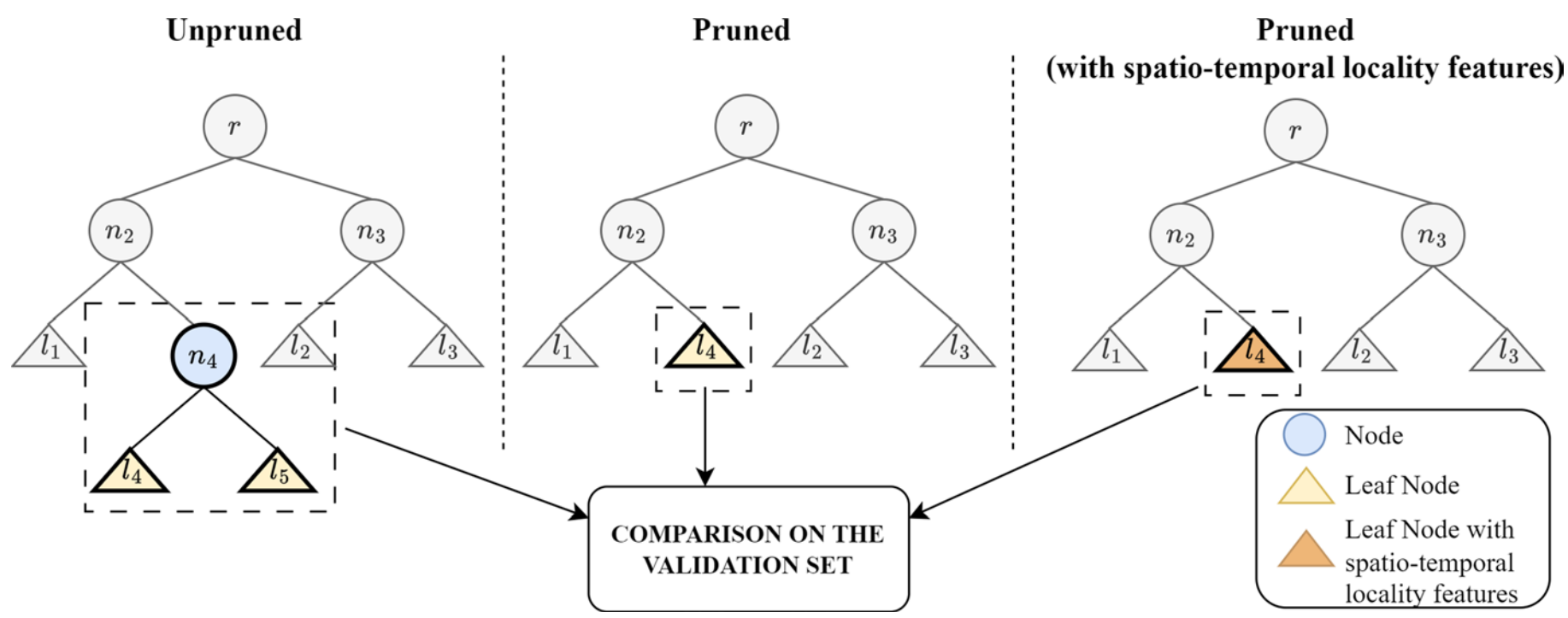

**Fig. 7** The proposed variant of the Reduced Error Pruning strategy that also takes into account the spatio-temporal locality

The time complexity of SPALT depends on: *i)* the construction of the initial linear model tree; *ii)* the introduction of the spatio-temporal locality into the leaf nodes; *iii)* the pruning phase. In the following, we analyze these steps separately.

While the final tree contains linear models only in the leaf nodes, we recall that the identification of the optimal split during the construction of the tree is based on the MSE achieved by the linear models learned for child nodes. Therefore, linear models are actually learned also for internal nodes, during the construction of the tree.

The identification of the optimal split for a node requires sorting the training instances according to the values of each input feature.[1] This step requires $O(m \cdot n \cdot \log n)$, where $n \cdot \log n$ is the complexity of optimal sorting algorithms. Then, for each possible threshold, a linear model tree is learned and evaluated, in terms of MSE. Since our implementation is based on quantile binning, a maximum number of $b$ thresholds is evaluated for each attribute. Therefore, $m \cdot b$ linear models are learned and evaluated. Considering that learning a linear model, and evaluating it in terms of MSE, takes[2] $O(m \cdot n)$, the asymptotic complexity to build a splitting node of the tree is $O(m \cdot n \cdot \log n) + O(m \cdot b) \cdot O(m \cdot n)$, that is $O(m \cdot n \cdot \log n) + O(m^2 \cdot n \cdot b)$. Moreover, considering that $b$ is a constant value (with $b \ll n$), the time complexity of building a node of the tree is $O(m \cdot n \cdot max(m, \log n))$.

In order to estimate the time complexity of building the whole linear model tree, we need to estimate the number of its nodes.

In the worst case, each leaf node will contain exactly $msp \cdot n$ training instances, leading to $\frac{n}{msp \cdot n} = \frac{1}{msp}$ leaf nodes. In this situation, the resulting tree will have $\sum_{i=0}^{\lfloor \log_2(1/msp) \rfloor} 2^i$ internal nodes, plus $\frac{1}{msp}$ leaf nodes. Therefore, in total, the number of nodes of the linear model tree is $O\left(\frac{1}{msp}\right)$.

In summary, the complexity of building the tree is:

$$O\left(\frac{1}{msp} \cdot m \cdot n \cdot max(m, \log n)\right) \tag{4}$$

The refinement of the linear models on the leaf nodes to consider the spatio-temporal locality requires building and evaluating an additional linear model for each leaf node. Before learning such models, it is necessary to perform a join operation between the instances fallen into the leaf node and all the other instances, related to the other locations fallen into the same leaf node, for the same time step (see Equation (2)). Considering that all the instances will fall into a leaf node, the join operation will overall take $O(n \cdot \log n)$, assuming that training instances are organized into a sorted data structure. An additional linear model is then learned and evaluated in terms of MSE on the validation set, for each leaf node. This takes $O(2 \cdot m \cdot n) = O(m \cdot n)$ in total,[3] for all the leaf nodes. It is finally worth noting that Equation (2) also requires the computation of the distance matrix $D$ (see also Equation (3)). However, since the number of distinct locations is much lower than the total number of instances $n$ and since it can be pre-computed once, its contribution to the time complexity can be considered negligible. In summary, the time complexity of this phase can be approximated to:

$$O(n \cdot \log n) + O(m \cdot n) = O(n \cdot max(m, \log n)) \tag{5}$$

[1] Sorting the training instances according to the values of the input feature is required only to identify the thresholds for numerical attributes.

[2] Note that the $n$ training instances are considered in the root only, while each internal node actually focuses on a subset.

[3] In this case, linear models are learned from $2m$ features (see Fig. 5).

Finally, the estimation of the time complexity of the pruning phase can be done by assuming the same cost of the previous phase, repeated for all the nodes of the tree, in the worst case scenario in which the tree is pruned up to the root node. Therefore, the worst-case time complexity of the pruning is:

$$O\left(\frac{1}{msp} \cdot n \cdot max(m, \log n)\right) \tag{6}$$

Summing up the time complexity of all the phases, namely, Equations (4), (5), and (6), we can conclude that SPALT asymptotically requires $O\left(\frac{1}{msp} \cdot m \cdot n \cdot max(m, \log n)\right)$, that corresponds to the time complexity of building a linear model tree (Equation (4)). This means that, although SPALT introduces additional computations to consider the spatio-temporal locality and to perform spatio-temporal pruning, it exhibits the same time complexity of learning a linear model tree that does not take into account the spatio-temporal locality and does not perform any pruning step.

## 4 Experiments

In this section, we describe the experiments that we performed to assess the effectiveness of SPALT. We run it with $b = 25$ (number of bins for the quantile binning) and $msp = 0.05$, namely, each leaf node should contain at least 5% of the training instances, to properly learn a linear model. In the experiments we considered different time horizons of the multi-step setting, namely, $h \in \{6, 12, 18\}$.

All the experiments were run on a server equipped with an Intel(R) Core i7-8700 CPU, 6 cores@3.2GHz, 64 GB of RAM and a NVIDIA GeForce RTX 3060 GPU.

As evaluation measures, we collected the Mean Absolute Error (MAE), the Root Mean Squared Error (RMSE) and the Relative Squared Error (RSE), averaged over the test instances and over the time steps of the respective multi-step setting. More formally, in the multi-step setting, they are defined as:

$$MAE = \frac{1}{h} \cdot \sum_{i=0}^{h} \frac{1}{|ts|} \cdot \sum_{y_{\beta,t} \in ts} |y_{\beta,t+i} - \widehat{y}_{\beta,t+i}| \tag{7}$$

$$RMSE = \frac{1}{h} \cdot \sum_{i=0}^{h} \sqrt{\frac{1}{|ts|} \cdot \sum_{y_{\beta,t} \in ts} (y_{\beta,t+i} - \widehat{y}_{\beta,t+i})^2} \tag{8}$$

$$RSE = \frac{1}{h} \cdot \sum_{i=0}^{h} \frac{\sum_{y_{\beta,t} \in ts} (y_{\beta,t+i} - \widehat{y}_{\beta,t+i})^2}{\sum_{y_{\beta,t} \in ts} (y_{\beta,t+i} - \overline{y}_{\beta,i})^2} \quad (9)$$

where *ts* is the test set, $y_{\beta,t}$ is the value of *y* at location $\beta$ and time *t* for a generic instance in *ts*, $\hat{y}_{\beta,t}$ is the value predicted for such an instance, and $\overline{y}_{\beta,i}$ is the average value of *y* at location $\beta$ for the *i*-th step ahead in the training set.

Note that, while MAE and RMSE provide an idea about the prediction errors made by the models, they are useful only for comparative purposes. On the contrary, the RSE directly measures the performance of the model with respect to a baseline represented by the mean value of *y* at a given location in the training set for the *i*-th step ahead (see the denominator of Equation (9)): an RSE value close to 0.0 means that the model perfectly predicts the value for all testing instances, while a value close to (respectively, higher than) 1.0 means that the model exhibits performances that are similar to (respectively, worse than) the prediction made by the mean.

In the following subsections, before reporting and discussing the obtained results, we describe the adopted datasets and the considered competitor approaches.

## 4.1 Datasets

We performed our experiments on three real datasets, for which we report quantitative information in Table 1:

**UKPV**: a public dataset consisting of four years of photovoltaic (PV) power production data from 1311 plants located in UK, recorded every 5 min. Missing values were imputed using the previous (or, when not available, the subsequent) value. Only plants having more than $50\%$ non-zero values were preserved, leading to a total of 332 plants. We randomly selected 8 non-consecutive days as test set; for each day of the test set, the preceding 4 days were considered as training set, and the last one was also used as validation set. Therefore, the whole dataset contains 8 $\times$ 478,080 instances, i.e., (382,464 training instances + 95,616 test instances) $\times$ 8.

**SDWPF** (Zhou et al., 2022): a dataset published in the context of the Baidu KDD Cup 2022 about the energy production collected for 245 days every 10 min from a wind farm having 134 turbines. This dataset was pre-processed following the approach adopted in Tan , Yue (2023),[4] mainly to fill in missing values. We randomly selected 8 weeks from the entire dataset as test set; for each week of the test set, the preceding 4 weeks were considered as training set, and the last one was also used as validation set. Therefore, the whole

**Table 1** Quantitative information of the considered datasets

| Dataset | Temporal resolution | Locations | Training interval | Testing interval | #training instances | #testing instances |
|---|---|---|---|---|---|---|
| UKPV | 5 min | 332 | 4 days | 1 day | 382,464 | 95,616 |
| SWDPF | 10 min | 134 | 4 weeks | 1 week | 540,288 | 135,072 |
| WPP | 15 min | 60 | 4 months | 1 month | 691,200 | 172,800 |

The number of training/testing instances refers to a single run out of all the runs performed by varying the test set dataset contains 675,360 × 8 instances, i.e., (540,288 training instances + 135,072 testing instances) × 8.

[4] https://github.com/LongxingTan/KDDCup2022-WPF

**WPP**: a real wind power plants dataset, provided by a lead company in the energy distribution field. The dataset consists of energy production data of 60 plants collected every 15 min for 1 year. All the months starting from the 5-th (8 in total) were alternatively considered as test set; for each month of the test set, the preceding 4 months were considered as training set, and the last one was also used as validation set. Therefore, the whole dataset contains 864,000 × 8 instances, i.e., (691,200 training instances + 172,800 test instances) × 8.

In all the datasets, the plants/turbines are associated with geographic coordinates (latitude and longitude), that enable the possibility to consider the spatial dimension. For all the datasets, we considered 12 historical measurements (i.e., $w = 12$) as input features. While SDWPF[5] and UKPV[6] are publicly available, WPP is not available because of commercial reasons. For reproducibility purposes, the code adopted to pre-process them and to partition data into training/validation/test sets is available at https://github.com/TinaDaversa/SPALT.

## 4.2 Competitors

We compared the performances achieved by SPALT with several competitor approaches able to work in the multi-step learning setting. In particular, we considered three different tree-based methods, namely, Regression Tree (henceforth denoted with **RT**), Random Forests (henceforth denoted with **RF**) and XGBoost (henceforth denoted with **XGB**). We adopted the implementation available in the python scikit-learn library v1.3.1, with their default parameter configuration.

Note that these approaches cannot naturally account for the spatial dimension. Therefore, we evaluated their performance by injecting spatial features into the dataset. These injected features were calculated using the PCNM method, as suggested in Ceci et al. (2017), to address spatial autocorrelation in methods that typically overlook it. These approaches will be denoted with **RT+PCNM**, **RF+PCNM**, and **XGB+PCNM**. For all these approaches, we also considered a variant in which data are preliminary differentiated to allow the learning methods to possibly better capture trends in the time series. Specifically, each value for a given time step is obtained by subtracting the preceding value of the time-series. We denote the variants based on this differentiation step with the **diff** suffix.

Finally, we considered four state-of-the-art neural network architectures that can naturally work in the multi-step setting and capture spatio-temporal phenomena. In particular, we considered **MTGNN** (Wu et al., 2020), **D$^2$STGNN** (Shao et al., 2022), **STAEformer** (Liu et al., 2023) and **HSTGNN** (Wang et al., 2025), with the parameter settings suggested in their respective papers. For such neural network architectures, the validation set was adopted to monitor the trend of the loss function and possibly trigger the early stopping criterion.

[5] **SDWPF dataset**: https://aistudio.baidu.com/competition/detail/152

[6] **UKPV dataset**: https://huggingface.co/datasets/openclimatefix/uk_pv

### 4.3 Results and discussion

In Tables 2, 3, and 4, we report all the results obtained by SPALT and by all the competitors, on all the datasets, for all the considered values of $h$, in terms of MAE, RMSE and RSE (mean and standard deviation). A quick look at the results emphasized in bold clearly shows that SPALT almost always outperforms all the considered competitor systems. In the few cases in which some competitors obtained the best performance, SPALT achieved comparable results, with an exception on the WPP dataset in terms of MAE and RMSE, where STAEformer provided significantly lower errors. However, in terms of RSE (see Table 4), the best performances on the WPP dataset were obtained by SPALT and HSTGNN, meaning that, in terms of actual usefulness of the predictions with respect to the baseline represented by the mean value, SPALT behaves better than STAEformer also on this dataset.

Going into the details of the results of tree-based competitors, it is clear that RT obtained the worst results, probably because regression trees are too simple to model the complex dynamics of the data at hand. The huge difference between RT and SPALT confirms the adequacy of adopting linear model trees in this context, and their effectiveness in capturing both linear and non-linear phenomena. On the other hand, RF and especially XGB obtained good performances, even if almost always behind SPALT. For all the tree-based competitors, the contribution coming from the injection of spatial features based on PCNM led to negligible differences, sometimes with a negative effect. This is possibly due to the fact that, contrary to the approach adopted in SPALT, PCNM features *statically* depend on the geographical locations of sensors and, therefore, cannot capture the spatio-temporal locality. Analogously, also the differentiation step did not provide significant differences, with some slight advantages provided to RF and XGB. However, they still fall behind SPALT. The negligible contribution of the differentiation step may be attributed to the lack of clear trends in the time series, since the primary purpose of differentiation is to mitigate such trends.

Looking at the results obtained by MTGNN and D$^2$STGNN, except for one case (D$^2$STGNN on the SDWPF dataset with $h = 6$), they surprisingly obtained the worst results among the considered systems. Moreover, in the UKPV dataset, the results obtained by D$^2$STGNN was also the worst in terms of standard deviation, showing that the results are not stable along the 8 randomly selected days. The more recent neural network architectures HSTGNN and STAEformer obtained generally worse performances on UKPV and SDWPF datasets, but were among the best-performing approaches on the WPP dataset. These results are possibly due to the complexity of these neural architectures that require a huge amount of training data to properly learn an accurate model. This is confirmed by the fact that, at least HSTGNN and STAEformer, obtained appreciable results on the WPP dataset, which is the largest among those considered in our experiments (see Table 1).

The results obtained with different time horizons show that higher values of $h$ lead to worse results, on average, with higher standard deviations. This is an expected behavior since it is more difficult to make predictions that are more distant in the future from the last observation. Focusing on the more challenging setting (i.e., $h = 18$), while some methods tend to achieve RSE values close to 1.0 (see RT, RT+PCNM, MTGNN and D$^2$STGNN, as well as HSTGNN and STAEformer on the SDWPF dataset), the RSE measured for SPALT remains clearly lower than 1.0, with a reasonably low standard deviation.

In order to depict a clear picture of the general performances obtainable by all the considered methods, in Figs. 8, 9, 10 we show their average rank, in terms of RMSE, over all the 24

**Table 2** MAE results (mean ± std dev) obtained by SPALT and the competitors, with different values of the prediction time horizon *h*

| | | MAE | | |
|---|---|---|---|---|
| | | h = 6 | h = 12 | h = 18 |
| UKPV dataset | SPALT | 8.29 ± 4.3 | **10.13 ± 4.9** | **11.67 ± 5.4** |
| | RT | 11.36 ± 5.6 | 13.85 ± 6.4 | 16.00 ± 7.0 |
| | RT diff | 11.87 ± 5.8 | 14.76 ± 6.7 | 17.37 ± 7.5 |
| | RT + PCNM | 11.52 ± 5.5 | 13.98 ± 6.3 | 16.00 ± 6.8 |
| | RT diff + PCNM | 12.04 ± 5.8 | 14.91 ± 6.6 | 17.49 ± 7.3 |
| | RF | 8.33 ± 4.3 | 10.26 ± 4.9 | 11.96 ± 5.4 |
| | RF diff | 8.22 ± 4.3 | 10.21 ± 4.9 | 12.04 ± 5.5 |
| | RF + PCNM | 8.43 ± 4.3 | 10.40 ± 4.8 | 12.11 ± 5.2 |
| | RF diff + PCNM | 8.28 ± 4.3 | 10.27 ± 4.9 | 12.11 ± 5.5 |
| | XGB | **8.28 ± 4.4** | 10.20 ± 5.0 | 11.88 ± 5.5 |
| | XGB diff | 8.32 ± 4.4 | 10.41 ± 5.2 | 12.30 ± 5.8 |
| | XGB + PCNM | 8.38 ± 4.4 | 10.36 ± 5.0 | 12.10 ± 5.5 |
| | XGB diff + PCNM | 8.32 ± 4.4 | 10.41 ± 5.2 | 12.30 ± 5.8 |
| | MTGNN | 29.08 ± 6.8 | 34.30 ± 10.1 | 34.07 ± 10.1 |
| | $D^2$STGNN | 22.18 ± 14.0 | 30.98 ± 15.8 | 34.08 ± 16.0 |
| | HSTGNN | 16.30 ± 11.6 | 22.10 ± 7.6 | 24.82 ± 9.6 |
| | STAEformer | 15.71 ± 12.2 | 17.13 ± 10.2 | 17.95 ± 7.8 |
| SDWPF dataset | SPALT | 95.6 ± 33.1 | **125.5 ± 40.1** | **147.6 ± 44.5** |
| | RT | 140.9 ± 47.6 | 180.8 ± 58.1 | 209.6 ± 64.9 |
| | RT diff | 142.0 ± 47.6 | 186.5 ± 59.3 | 218.4 ± 66.6 |
| | RT + PCNM | 143.6 ± 48.1 | 183.5 ± 58.6 | 212.2 ± 65.5 |
| | RT diff + PCNM | 142.4 ± 48.0 | 186.7 ± 59.31 | 218.9 ± 66.5 |
| | RF | 102.5 ± 35.1 | 134.5 ± 42.8 | 157.8 ± 47.5 |
| | RF diff | 98.6 ± 32.9 | 130.5 ± 40.1 | 153.8 ± 44.4 |
| | RF + PCNM | 102.6 ± 35.1 | 134.6 ± 42.7 | 157.4 ± 47.4 |
| | RF diff + PCNM | 98.9 ± 33.1 | 130.9 ± 40.1 | 154.2 ± 44.0 |
| | XGB | 98.6 ± 33.9 | 129.6 ± 41.4 | 151.5 ± 46.0 |
| | XGB diff | 97.1 ± 32.4 | 128.4 ± 39.3 | 150.8 ± 43.3 |
| | XGB + PCNM | 98.6 ± 33.9 | 129.5 ± 41.2 | 151.3 ± 45.8 |
| | XGB diff + PCNM | 110.6 ± 37.6 | 134.3 ± 26.8 | 138.6 ± 47.8 |
| | MTGNN | 121.4 ± 27.8 | 197.7 ± 37.0 | 223.6 ± 44.2 |
| | $D^2$STGNN | **89.3 ± 32.2** | 154.9 ± 94.2 | 200.0 ± 24.0 |
| | HSTGNN | 121.3 ± 36.6 | 163.3 ± 45.3 | 194.0 ± 56.1 |
| | STAEformer | 128.3 ± 46.8 | 167.3 ± 55.4 | 197.8 ± 63.5 |
| WPP dataset | SPALT | 0.357 ± 0.09 | 0.489 ± 0.11 | 0.586 ± 0.12 |
| | RT | 0.523 ± 0.12 | 0.707 ± 0.15 | 0.836 ± 0.17 |
| | RT diff | 0.544 ± 0.12 | 0.746 ± 0.15 | 0.893 ± 0.17 |
| | RT + PCNM | 0.524 ± 0.12 | 0.703 ± 0.15 | 0.828 ± 0.16 |
| | RT diff + PCNM | 0.544 ± 0.12 | 0.745 ± 0.15 | 0.890 ± 0.17 |
| | RF | 0.379 ± 0.09 | 0.521 ± 0.11 | 0.626 ± 0.12 |
| | RF diff | 0.372 ± 0.09 | 0.514 ± 0.10 | 0.619 ± 0.12 |
| | RF + PCNM | 0.379 ± 0.09 | 0.521 ± 0.11 | 0.624 ± 0.12 |
| | RF diff + PCNM | 0.371 ± 0.09 | 0.513 ± 0.10 | 0.616 ± 0.12 |
| | XGB | 0.371 ± 0.09 | 0.509 ± 0.11 | 0.609 ± 0.12 |
| | XGB diff | 0.364 ± 0.09 | 0.502 ± 0.10 | 0.603 ± 0.12 |
| | XGB + PCNM | 0.371 ± 0.09 | 0.509 ± 0.11 | 0.609 ± 0.12 |

**Table 2** (continued)

| | | | | |
|---|---|---|---|---|
| | XGB diff + PCNM | 0.363 ± 0.08 | 0.501 ± 0.10 | 0.600 ± 0.12 |
| | MTGNN | 0.550 ± 0.11 | 0.892 ± 0.12 | 0.997 ± 0.13 |
| | $D^2$STGNN | 0.624 ± 0.22 | 0.631 ± 0.22 | 0.885 ± 0.22 |
| | HSTGNN | 0.355 ± 0.09 | 0.491 ± 0.13 | 0.588 ± 0.15 |
| | STAEformer | **0.268 ± 0.05** | **0.328 ± 0.07** | **0.461 ± 0.09** |

The best result, for each value of $h$ and each measure, is emphasized in bold performed runs (3 datasets $\times$ 8 runs for each dataset), with the three considered values of $h$. The figures show the superiority of SPALT for all the values of $h$, followed by XGB+PCNM when $h = 6$, and by XGB when $h = 12$ or $h = 18$. Notably, the difference in terms of average rank is close to 3 when $h = 6$ and close to 2 when $h = 12$ or $h = 18$, meaning that the advantage of SPALT appears significant. In order to confirm such an advantage from a statistical viewpoint, we conducted a set of Wilcoxon signed-rank tests between SPALT and XGB, and between SPALT and XGB+PCNM, with all the values of $h$. The obtained $p$-values, corrected according to the False Discovery Rate (FDR) for multiple tests proposed by Benjamini and Hochberg (1995), are reported in Table 5. As we can observe, the difference between SPALT and XGB as well as the difference between SPALT and XGB+PCNM is statistically significant with $\alpha = 0.001$, for all the considered values of $h$.

We also collected the average running times of all the considered learning algorithms, which are reported in Table 6, for each considered dataset and value of $h$. As it can be observed, SPALT requires a running time that is always lower than that required by RF and all the considered neural network architectures. Only the methods based on RT and XGB exhibited lower running times. However, their performances in terms of predictive errors, as already discussed, are significantly behind those of SPALT. A final observation can be drawn in terms of scalability of SPALT: running times exhibited by SPALT appears to almost linearly depend on the number of instances, which is coherent with analysis of its computational complexity (see Sect. 3.5).

As regards the exploitation of the spatial dimension, the approach adopted by SPALT to capture the spatio-temporal locality appeared to be effective, and led to the best results. In order to assess the specific contribution provided by the consideration of the spatio-temporal locality, as well as of the proposed pruning approach (that also considers the spatio-temporal locality while evaluating whether to prune a sub-tree or not), we conducted an ablation study. In particular, we ran the experiments with a variant of SPALT that does not perform the pruning step (henceforth called **SPALT-NP**) and with a variant of SPALT that neither performs the pruning step nor considers the spatio-temporal locality in the leaf nodes, which is analogous to a classical linear model tree able to perform multi-step predictions (henceforth called **SPALT-NP-NS**). While the comparison with the first variant allows us to evaluate the contribution provided by the pruning phase (both in terms of prediction errors and in terms of complexity of the model), the comparison with the second variant specifically allows us to assess the contribution of considering the spatio-temporal locality. The results of the ablation study are reported in Table 7, in terms of reduction of MAE, RMSE, RSE and number of nodes, achieved by SPALT over the considered variants. Higher (positive) values indicate greater improvement obtained by SPALT.

Focusing on SPALT-NP, we can observe that the proposed pruning step may exhibit a positive influence (see the results with $h = 12$ and $h = 18$ on SDWPF), but also a negligible (see the results on WPP) or even negative influence (see the results on UKPV, with $h = 12$

**Table 3** RMSE results (mean ± std dev) obtained by SPALT and the competitors, with different values of the prediction time horizon *h*

| | | RMSE | | |
|---|---|---|---|---|
| | | h = 6 | h = 12 | h = 18 |
| UKPV dataset | SPALT | **18.35 ± 8.4** | **20.83 ± 9.0** | **22.66 ± 9.4** |
| | RT | 26.71 ± 11.1 | 30.57 ± 11.6 | 33.76 ± 12.0 |
| | RT diff | 27.74 ± 11.5 | 32.18 ± 12.2 | 35.94 ± 12.8 |
| | RT + PCNM | 26.98 ± 10.8 | 30.96 ± 11.3 | 34.09 ± 11.7 |
| | RT diff + PCNM | 28.07 ± 11.5 | 32.47 ± 12.1 | 36.23 ± 12.6 |
| | RF | 18.94 ± 8.4 | 21.72 ± 8.9 | 24.00 ± 9.3 |
| | RF diff | 19.21 ± 8.7 | 22.15 ± 9.3 | 24.67 ± 9.9 |
| | RF + PCNM | 19.06 ± 8.3 | 22.04 ± 8.7 | 24.48 ± 9.0 |
| | RF diff+ PCNM | 19.32 ± 8.8 | 22.31 ± 9.3 | 24.87 ± 9.9 |
| | XGB | 18.88 ± 8.5 | 21.63 ± 9.1 | 23.88 ± 9.5 |
| | XGB diff | 19.58 ± 8.9 | 22.80 ± 9.8 | 25.46 ± 10.6 |
| | XGB + PCNM | 19.00 ± 8.5 | 21.86 ± 9.0 | 24.22 ± 9.5 |
| | XGB diff + PCNM | 19.59 ± 8.9 | 22.80 ± 9.8 | 25.47 ± 10.6 |
| | MTGNN | 41.92 ± 9.4 | 47.55 ± 10.9 | 47.62 ± 10.9 |
| | $D^2$STGNN | 30.04 ± 15.8 | 39.66 ± 16.3 | 43.62 ± 17.3 |
| | HSTGNN | 26.44 ± 12.6 | 30.50 ± 9.5 | 38.05 ± 14.8 |
| | STAEformer | 26.12 ± 17.2 | 29.14 ± 15.6 | 31.48 ± 11.9 |
| SDWPF dataset | SPALT | 152.5 ± 43.9 | **190.7 ± 51.0** | **217.5 ± 55.0** |
| | RT | 227.7 ± 62.4 | 280.2 ± 73.0 | 316.7 ± 79.4 |
| | RT diff | 232.5 ± 61.8 | 291.5 ± 74.9 | 332.0 ± 82.2 |
| | RT + PCNM | 230.4 ± 62.5 | 283.4 ± 73.3 | 319.4 ± 79.9 |
| | RT diff + PCNM | 232.4 ± 62.7 | 290.7 ± 75.0 | 331.5 ± 82.0 |
| | RF | 163.4 ± 45.2 | 203.6 ± 53.4 | 230.7 ± 57.6 |
| | RF diff | 161.5 ± 44.6 | 203.3 ± 53.6 | 232.2 ± 58.6 |
| | RF + PCNM | 162.9 ± 45.1 | 203.0 ± 53.1 | 229.9 ± 57.3 |
| | RF diff + PCNM | 161.1 ± 44.6 | 202.5 ± 53.3 | 231.1 ± 57.7 |
| | XGB | 159.2 ± 44.3 | 198.1 ± 52.2 | 224.2 ± 56.4 |
| | XGB diff | 159.6 ± 44.3 | 200.6 ± 53.1 | 228.5 ± 57.6 |
| | XGB + PCNM | 159.1 ± 44.3 | 197.9 ± 52.1 | 223.8 ± 56.1 |
| | XGB diff + PCNM | 166.3 ± 42.7 | 202.7 ± 67.9 | 226.0 ± 59.5 |
| | MTGNN | 181.9 ± 40.6 | 261.1 ± 46.0 | 288.2 ± 53.8 |
| | $D^2$STGNN | **151.1 ± 43.9** | 223.5 ± 96.1 | 287.7 ± 37.4 |
| | HSTGNN | 179.5 ± 49.2 | 235.4 ± 60.7 | 276.0 ± 72.7 |
| | STAEformer | 208.8 ± 61.1 | 257.4 ± 68.5 | 298.3 ± 77.7 |
| WPP dataset | SPALT | 0.657 ± 0.14 | 0.850 ± 0.16 | 0.985 ± 0.18 |
| | RT | 0.961 ± 0.18 | 1.249 ± 0.21 | 1.442 ± 0.23 |
| | RT diff | 0.990 ± 0.18 | 1.300 ± 0.21 | 1.516 ± 0.23 |
| | RT + PCNM | 0.964 ± 0.18 | 1.244 ± 0.20 | 1.432 ± 0.22 |
| | RT diff + PCNM | 0.990 ± 0.18 | 1.297 ± 0.20 | 1.512 ± 0.23 |
| | RF | 0.687 ± 0.14 | 0.891 ± 0.16 | 1.031 ± 0.18 |
| | RF diff | 0.687 ± 0.14 | 0.902 ± 0.17 | 1.051 ± 0.19 |
| | RF + PCNM | 0.684 ± 0.14 | 0.889 ± 0.16 | 1.029 ± 0.18 |
| | RF diff+ PCNM | 0.684 ± 0.14 | 0.897 ± 0.17 | 1.046 ± 0.19 |
| | XGB | 0.683 ± 0.15 | 0.883 ± 0.17 | 1.019 ± 0.19 |
| | XGB diff | 0.681 ± 0.14 | 0.894 ± 0.17 | 1.041 ± 0.20 |
| | XGB + PCNM | 0.679 ± 0.15 | 0.879 ± 0.17 | 1.014 ± 0.19 |

**Table 3** (continued)

| | | | | |
|---|---|---|---|---|
| | XGB diff + PCNM | 0.677 ± 0.14 | 0.887 ± 0.17 | 1.033 ± 0.19 |
| | MTGNN | 0.848 ± 0.23 | 1.315 ± 0.20 | 1.450 ± 0.22 |
| | $D^2$STGNN | 0.938 ± 0.24 | 1.044 ± 0.18 | 1.372 ± 0.24 |
| | HSTGNN | 0.656 ± 0.14 | 0.869 ± 0.19 | 1.020 ± 0.22 |
| | STAEformer | **0.468 ± 0.08** | **0.566 ± 0.10** | **0.774 ± 0.13** |

The best result, for each value of $h$ and each measure, is emphasized in bold and $h = 18$). On the other hand, the trees obtained by SPALT show a drop ranging from about 23% (see the SDWPF and WPP datasets, with $h = 6$) to about 35% (see the UKPV dataset with $h = 6$ and the SDWPF dataset with $h = 12$) in the number of nodes, making the models more compact and interpretable.

If we ignore the spatio-temporal locality (see the lines of SPALT-NP-NS in Table 7), we can see that the improvement provided by SPALT is present for all the considered measures, for all the values of $h$ and for all the datasets. In order to statistically evaluate if such an advantage is significant, we conducted a set of Wilcoxon signed-rank tests, by performing a correction according to the False Discovery Rate (FDR) for multiple tests proposed by Benjamini and Hochberg (1995). The obtained $p$-values are reported in Table 8. As we can observe, the difference between SPALT and SPALT-NP-NS is statistically significant with $\alpha = 0.001$, for all the evaluation measures including the number of nodes. Therefore, the approach we propose to consider the spatio-temporal locality systematically provides advantages in terms of predictive accuracy, since it selectively introduces spatio-temporal locality features when considered beneficial. In Table 10, we show the percentage of leaf nodes in which spatio-temporal locality features were injected by SPALT for each dataset and for each value of $h$. From the table it is clear that, for several leaf nodes (roughly, between 64% and 77%) SPALT considered the introduction of spatio-temporal locality features beneficial for the performance of the corresponding linear model on the validation set.

In Fig. 11 we report an example of a tree learned by SPALT from the UKPV dataset ($h$=18, first run). As we can see, some leaf nodes exploit spatio-temporal locality features (STLfeatures: True), while others do not exploit them (STLfeatures: False). In Figs. 12a and 12b, we plot the time series of instances fallen into two representative nodes of such a tree: node 7, which exploits spatio-temporal locality features, and node 5 that does not exploit them. From the figures, we can observe a low dispersion and a clear pattern for time series in node 5, while time series in node 7 do not show an equally clear pattern. This difference possibly motivates the exploitation of spatio-temporal locality features in node 7, which led to a reduction of errors on the validation set. This hypothesis is further confirmed by the distribution of locations of instances fallen into nodes 5 and 7, shown in Figs. 12c and 12d, respectively. Looking at Fig. 12c, it is clear that instances falling into node 5 are distributed quite evenly across different locations, with a mean location that almost corresponds to that at node 0 (root of the tree, which includes all the instances of the dataset). On the other hand, instances falling into node 7 (see Fig. 12d) mostly belong to a few locations with low latitude values, with a mean location which is clearly distinct from the mean at node 0. This confirms the fact that, in this case, features coming from instances at nearby locations clustered in the same leaf node may contribute to the reduction of the prediction errors.

The difference introduced by the pruning (alone) is not significant, in terms of MAE, RMSE and RSE, while it is significant with $\alpha = 0.001$ in terms of the number of nodes. This result confirms that the proposed pruning strategy, while not consistently improving

**Table 4** RSE results (mean ± std dev) obtained by SPALT and the competitors, with different values of the prediction time horizon *h*

| | | RSE | | |
|---|---|---|---|---|
| | | h = 6 | h = 12 | h = 18 |
| UKPV dataset | SPALT | **0.145 ± 0.07** | **0.187 ± 0.09** | **0.220 ± 0.09** |
| | RT | 0.311 ± 0.12 | 0.413 ± 0.14 | 0.507 ± 0.16 |
| | RT diff | 0.340 ± 0.14 | 0.465 ± 0.16 | 0.587 ± 0.19 |
| | RT + PCNM | 0.320 ± 0.12 | 0.430 ± 0.15 | 0.529 ± 0.17 |
| | RT diff + PCNM | 0.35 ± 0.14 | 0.475 ± 0.17 | 0.602 ± 0.20 |
| | RF | 0.155 ± 0.07 | 0.204 ± 0.08 | 0.248 ± 0.09 |
| | RF diff | 0.160 ± 0.08 | 0.212 ± 0.09 | 0.261 ± 0.09 |
| | RF + PCNM | 0.158 ± 0.07 | 0.215 ± 0.09 | 0.268 ± 0.10 |
| | RF diff + PCNM | 0.161 ± 0.08 | 0.215 ± 0.09 | 0.266 ± 0.09 |
| | XGB | 0.154 ± 0.08 | 0.201 ± 0.09 | 0.243 ± 0.09 |
| | XGB diff | 0.166 ± 0.08 | 0.223 ± 0.10 | 0.276 ± 0.11 |
| | XGB + PCNM | 0.156 ± 0.08 | 0.207 ± 0.09 | 0.253 ± 0.09 |
| | XGB diff + PCNM | 0.166 ± 0.08 | 0.223 ± 0.10 | 0.276 ± 0.11 |
| | MTGNN | 0.722 ± 0.05 | 0.866 ± 0.02 | 0.871 ± 0.02 |
| | $D^2$STGNN | 0.476 ± 0.39 | 0.693 ± 0.49 | 0.821 ± 0.35 |
| | HSTGNN | 0.241 ± 0.25 | 0.462 ± 0.17 | 0.663 ± 0.21 |
| | STAEformer | 0.240 ± 0.23 | 0.408 ± 0.17 | 0.441 ± 0.20 |
| SDWPF dataset | SPALT | 0.213 ± 0.10 | **0.344 ± 0.14** | **0.457 ± 0.18** |
| | RT | 0.474 ± 0.21 | 0.735 ± 0.31 | 0.956 ± 0.39 |
| | RT diff | 0.495 ± 0.21 | 0.802 ± 0.33 | 1.061 ± 0.42 |
| | RT + PCNM | 0.487 ± 0.22 | 0.754 ± 0.32 | 0.973 ± 0.40 |
| | RT diff + PCNM | 0.494 ± 0.21 | 0.796 ± 0.33 | 1.057 ± 0.42 |
| | RF | 0.245 ± 0.11 | 0.392 ± 0.17 | 0.513 ± 0.21 |
| | RF diff | 0.24 ± 0.10 | 0.394 ± 0.17 | 0.525 ± 0.22 |
| | RF + PCNM | 0.244 ± 0.11 | 0.389 ± 0.17 | 0.508 ± 0.21 |
| | RF diff + PCNM | 0.238 ± 0.11 | 0.39 ± 0.17 | 0.519 ± 0.22 |
| | XGB | 0.232 ± 0.10 | 0.370 ± 0.16 | 0.482 ± 0.19 |
| | XGB diff | 0.235 ± 0.10 | 0.383 ± 0.17 | 0.507 ± 0.21 |
| | XGB + PCNM | 0.232 ± 0.10 | 0.369 ± 0.16 | 0.480 ± 0.19 |
| | XGB diff + PCNM | 0.272 ± 0.05 | 0.392 ± 0.19 | 0.459 ± 0.40 |
| | MTGNN | 0.349 ± 0.07 | 0.675 ± 0.07 | 0.793 ± 0.10 |
| | $D^2$STGNN | **0.211 ± 0.09** | 0.450 ± 0.27 | 0.713 ± 0.20 |
| | HSTGNN | 0.303 ± 0.14 | 0.536 ± 0.24 | 0.749 ± 0.34 |
| | STAEformer | 0.295 ± 0.14 | 0.463 ± 0.19 | 0.677 ± 0.23 |
| WPP dataset | SPALT | 0.145 ± 0.02 | **0.248 ± 0.04** | **0.334 ± 0.06** |
| | RT | 0.315 ± 0.04 | 0.539 ± 0.08 | 0.718 ± 0.12 |
| | RT diff | 0.336 ± 0.04 | 0.586 ± 0.10 | 0.802 ± 0.15 |
| | RT + PCNM | 0.316 ± 0.04 | 0.533 ± 0.08 | 0.708 ± 0.12 |
| | RT diff + PCNM | 0.336 ± 0.04 | 0.586 ± 0.10 | 0.799 ± 0.16 |
| | RF | 0.160 ± 0.02 | 0.273 ± 0.04 | 0.366 ± 0.06 |
| | RF diff | 0.16 ± 0.02 | 0.28 ± 0.05 | 0.383 ± 0.08 |
| | RF + PCNM | 0.158 ± 0.02 | 0.271 ± 0.04 | 0.363 ± 0.06 |
| | RF diff + PCNM | 0.158 ± 0.02 | 0.277 ± 0.04 | 0.379 ± 0.07 |
| | XGB | 0.156 ± 0.02 | 0.266 ± 0.04 | 0.355 ± 0.06 |
| | XGB diff | 0.158 ± 0.02 | 0.275 ± 0.05 | 0.375 ± 0.07 |
| | XGB + PCNM | 0.155 ± 0.02 | 0.262 ± 0.03 | 0.350 ± 0.05 |

**Table 4** (continued)

| | | | | |
|---|---|---|---|---|
| | XGB diff + PCNM | 0.155 ± 0.01 | 0.270 ± 0.04 | 0.368 ± 0.07 |
| | MTGNN | 0.535 ± 0.36 | 0.722 ± 0.04 | 0.826 ± 0.05 |
| | D$^2$STGNN | 0.350 ± 0.21 | 0.401 ± 0.17 | 0.687 ± 0.21 |
| | HSTGNN | **0.142 ± 0.01** | 0.252 ± 0.02 | 0.347 ± 0.03 |
| | STAEformer | 0.179 ± 0.03 | 0.290 ± 0.06 | 0.400 ± 0.08 |

The best result, for each value of $h$ and each measure, is emphasized in bold

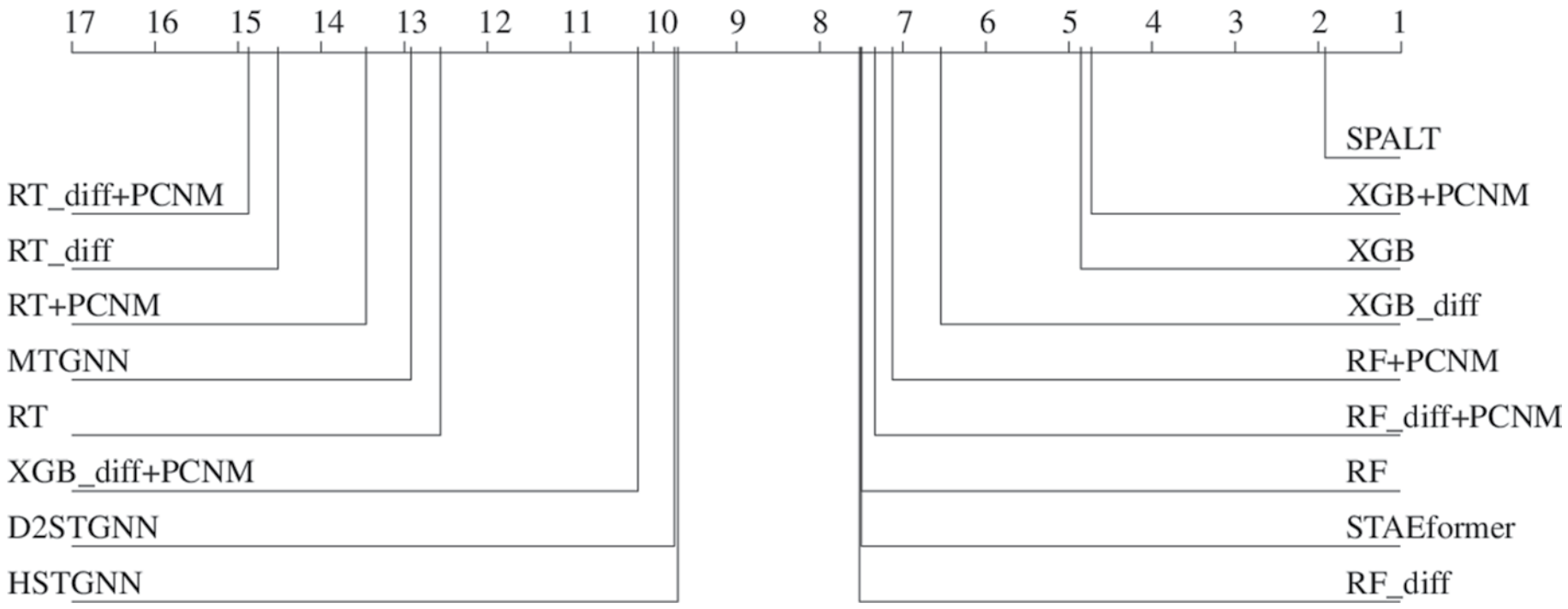

**Fig. 8** Average ranks computed on the RMSE results on all datasets, with $h = 6$

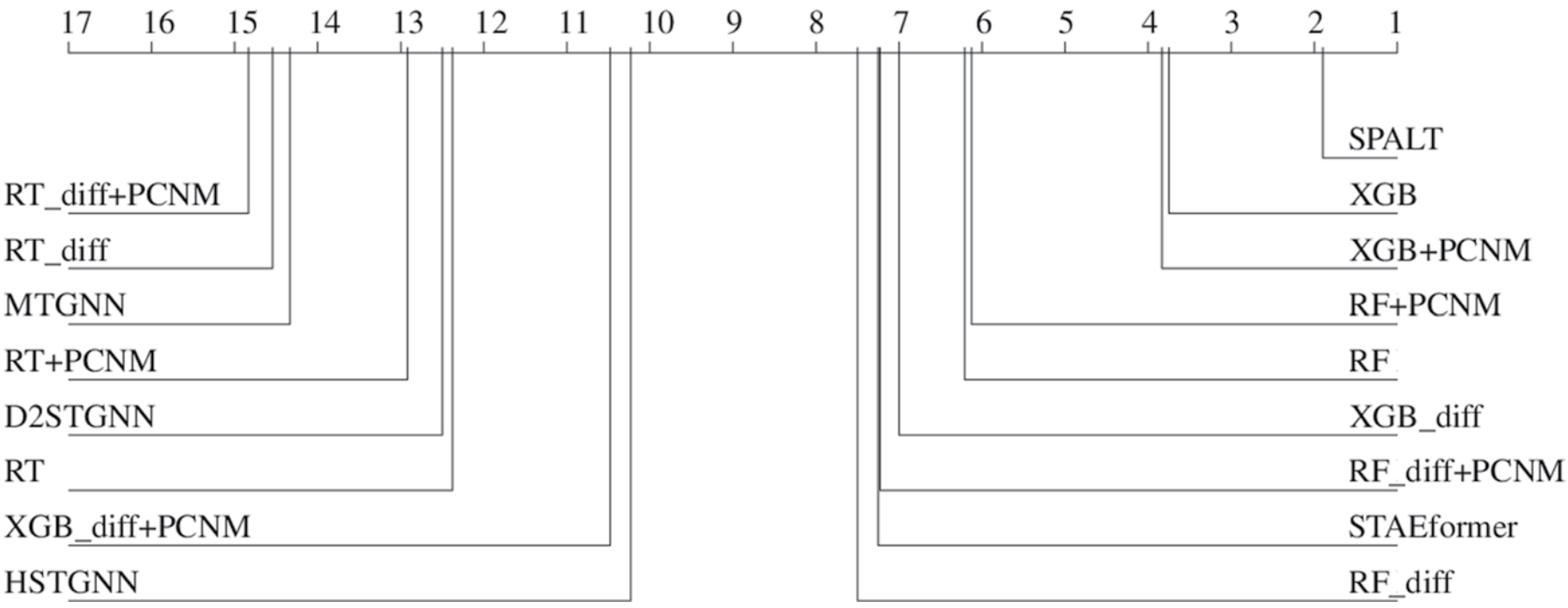

**Fig. 9** Average ranks computed on the RMSE results on all datasets, with $h = 12$

MAE, RMSE, and RSE (as shown in Table 7), is generally preferable because it results in significantly less complex models. Moreover, as can be observed in Table 9, the pruning phase does not significantly affect the overall running times. Therefore, its adoption can generally be considered beneficial.

A final analysis that we conducted regards the possible influence of some design choices within SPALT. Specifically, we analyze the possible influence of the linear weighting we adopt with respect to the distances, when injecting spatio-temporal locality features (see Equations (2) and (3)). In this respect, we ran additional experiments on the intermediate horizon ($h = 12$) considering a quadratic weighting of the distances, leading to a variant of the Equation (3), as follows:

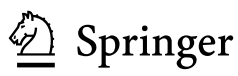

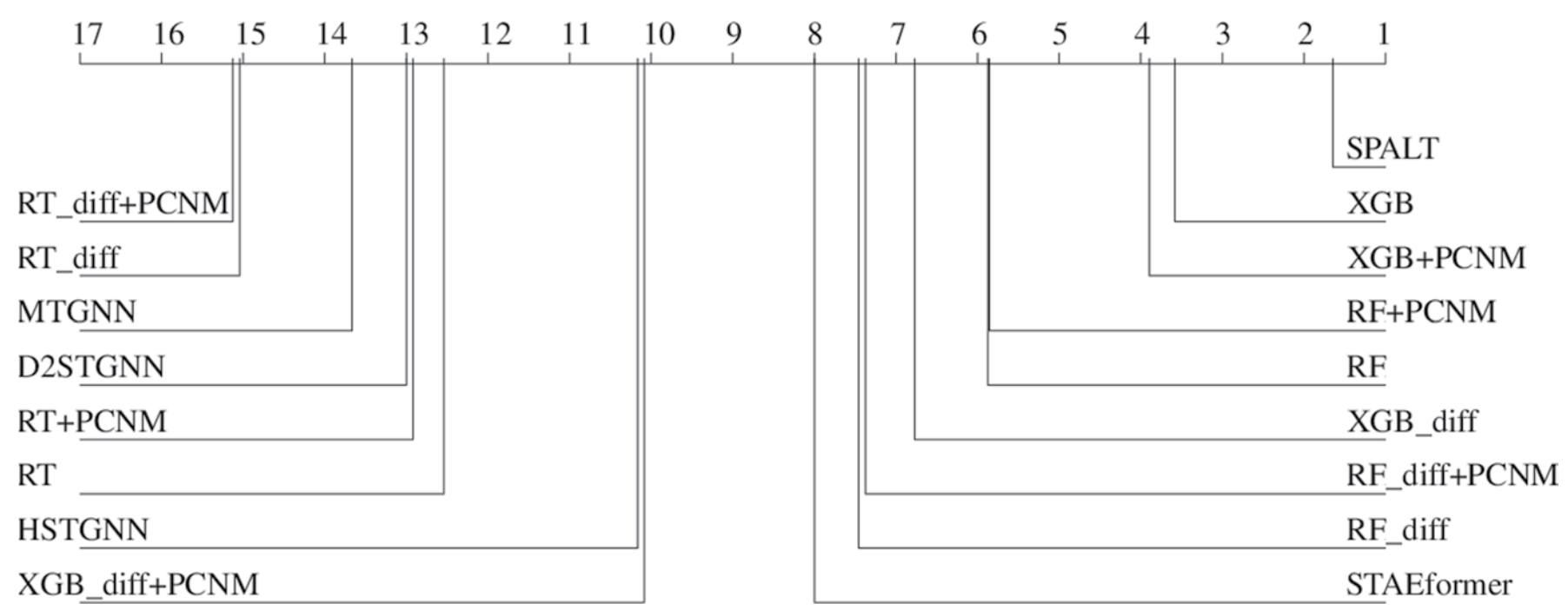

**Fig. 10** Average ranks computed on the RMSE results on all datasets, with $h = 18$

**Table 5** P-values obtained by the Wilcoxon signed-rank test with the False Discovery Rate (FDR) correction, computed on the RMSE, between SPALT and its closest competitors (according to the average ranks on the RMSE - see Figs. 8, 9 and 10)

| | h = 6 | h = 12 | h = 18 |
|---|---|---|---|
| XGB | $< 0.001$ | $< 0.001$ | $< 0.001$ |
| XGB+PCNM | $< 0.001$ | $< 0.001$ | $< 0.001$ |

**Table 6** Average running times (s) measured for SPALT and the competitors, with different values of the prediction time horizon $h$

| | *UKPV dataset* | | | *SDWPF dataset* | | | *WPP dataset* | | |
|---|---|---|---|---|---|---|---|---|---|
| *Time (s)* | h=6 | h=12 | h=18 | h=6 | h=12 | h=18 | h=6 | h=12 | h=18 |
| SPALT | 169.40 | 170.66 | 173.58 | 168.72 | 209.50 | 242.25 | 176.72 | 210.98 | 249.49 |
| RT | 30.19 | 29.82 | 31.53 | 55.33 | 58.22 | 60.68 | 68.05 | 74.44 | 82.81 |
| RT diff | 55.01 | 71.06 | 81.63 | 60.33 | 72.86 | 88.66 | 105.61 | 135.10 | 159.84 |
| RT + PCNM | 46.28 | 46.44 | 48.55 | 76.68 | 82.48 | 91.88 | 71.37 | 75.80 | 82.87 |
| RT diff + PCNM | 125.53 | 168.60 | 201.94 | 91.63 | 112.07 | 133.56 | 139.57 | 155.13 | 168.85 |
| RF | 274.36 | 274.02 | 293.62 | 520.78 | 534.86 | 561.66 | 602.06 | 669.41 | 732.65 |
| RF diff | 326.62 | 407.69 | 483.20 | 411.60 | 473.23 | 564.30 | 631.75 | 698.35 | 720.33 |
| RF + PCNM | 277.25 | 282.21 | 309.04 | 453.81 | 496.51 | 551.90 | 427.67 | 459.34 | 499.56 |
| RF diff + PCNM | 386.81 | 459.29 | 543.88 | 545.71 | 669.32 | 813.88 | 665.88 | 708.05 | 790.45 |
| XGB | 4.09 | 8.91 | 14.69 | 8.11 | 11.30 | 18.83 | 7.48 | 15.25 | 30.41 |
| XGB diff | 4.74 | 10.46 | 17.13 | 4.15 | 9.79 | 16.37 | 4.82 | 10.75 | 16.59 |
| XGB + PCNM | 4.86 | 10.84 | 17.60 | 8.89 | 13.52 | 21.36 | 7.51 | 16.02 | 31.99 |
| XGB diff + PCNM | 4.72 | 10.51 | 17.21 | 7.77 | 16.10 | 25.50 | 6.62 | 14.87 | 22.88 |
| MTGNN | 359.46 | 361.00 | 363.24 | 498.15 | 498.00 | 501.78 | 740.23 | 732.77 | 734.36 |
| D2STGNN | 499.55 | 500.83 | 502.70 | 615.12 | 615.00 | 618.15 | 816.86 | 810.64 | 811.97 |
| HSTGNN | 839.46 | 841.00 | 843.24 | 978.15 | 978.00 | 981.78 | 1220.23 | 1212.77 | 1214.36 |
| STAEFormer | 1776.48 | 1776.50 | 1776.46 | 2973.84 | 2033.07 | 2033.91 | 2400.44 | 2351.21 | 2391.41 |

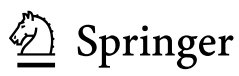

**Table 7** Ablation analysis: reduction of errors and of the number of nodes achieved by SPALT with respect to its variants without pruning (SPALT-NP), and without pruning and spatio-temporal locality (SPALT-NP-NS)

| | MAE | | | RMSE | | | RSE | | | # nodes | | |
|---|---|---|---|---|---|---|---|---|---|---|---|---|
| | h=6 | h=12 | h=18 | h=6 | h=12 | h=18 | h=6 | h=12 | h=18 | h=6 | h=12 | h=18 |
| *UKPV dataset* | | | | | | | | | | | | |
| SPALT-NP | −0.6% | −1.6% | −1.8% | −0.1% | −1.0% | −1.0% | 0.0% | −2.1% | −2.3% | 35.2% | 30.4% | 28.6% |
| SPALT-NP-NS | 3.5% | 5.9% | 8.3% | 3.9% | 5.9% | 8.2% | 7.6% | 11.4% | 15.7% | 35.2% | 30.4% | 28.6% |
| *SDWPF dataset* | | | | | | | | | | | | |
| SPALT-NP | −0.3% | 0.3% | 0.0% | −0.4% | 0.2% | 0.1% | −0.9% | 0.6% | 0.2% | 23.8% | 35.6% | 31.4% |
| SPALT-NP-NS | 3.1% | 3.1% | 2.0% | 3.6% | 3.2% | 2.2% | 7.0% | 6.0% | 3.6% | 23.8% | 35.6% | 31.4% |
| *WPP dataset* | | | | | | | | | | | | |
| SPALT-NP | 0.0% | 0.0% | 0.0% | −0.3% | −0.1% | −0.1% | 0.0% | −0.4% | −0.3% | 23.4% | 30.0% | 24.3% |
| SPALT-NP-NS | 3.0% | 3.7% | 3.8% | 2.2% | 2.9% | 2.8% | 5.2% | 5.7% | 5.4% | 23.4% | 30.0% | 24.3% |

**Table 8** Ablation analysis: p-values of the Wilcoxon signed-rank test with the False Discovery Rate (FDR) correction, between SPALT and its variants without pruning (SPALT-NP), and without pruning and spatio-temporal locality (SPALT-NP-NS)

| | MAE | | | RMSE | | | RSE | | | # nodes | | |
|---|---|---|---|---|---|---|---|---|---|---|---|---|
| | h=6 | h=12 | h=18 | h=6 | h=12 | h=18 | h=6 | h=12 | h=18 | h=6 | h=12 | h=18 |
| SPALT-NP | 0.126 | 0.501 | 0.417 | 0.994 | 1.000 | 1.000 | 0.752 | 0.808 | 0.590 | **<0.001** | **<0.001** | **<0.001** |
| SPALT-NP-NS | **<0.001** | **<0.001** | **<0.001** | **<0.001** | **<0.001** | **<0.001** | **<0.001** | **<0.001** | **<0.001** | **<0.001** | **<0.001** | **<0.001** |

Statistically significant results are reported in bold

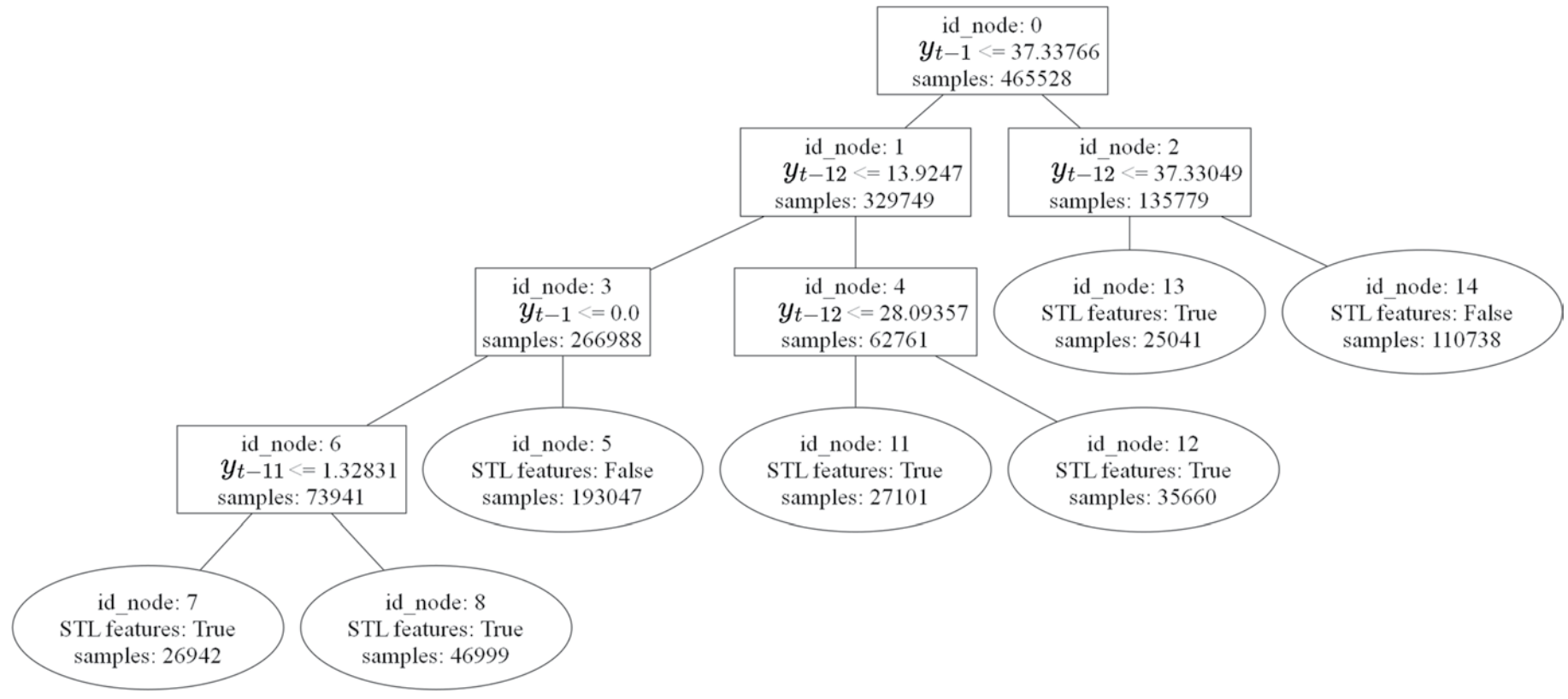

**Fig. 11** An example of a tree learned by SPALT (UKPV dataset, $h$=18, first run)

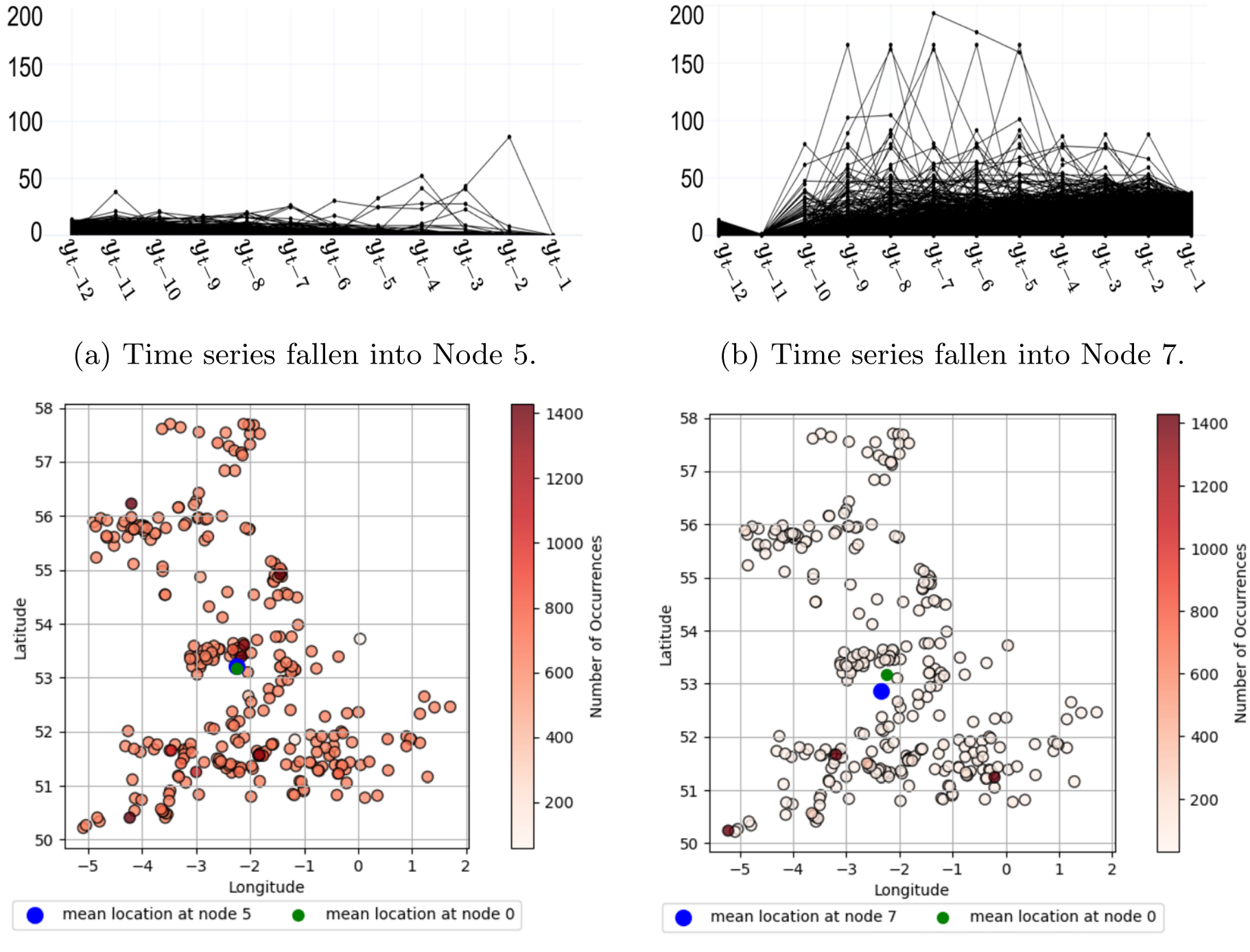

(a) Time series fallen into Node 5.

(b) Time series fallen into Node 7.

(c) Distribution of locations of the instances fallen into Node 5.

(d) Distribution of locations of the instances fallen into Node 7.

**Fig. 12** Analysis of the time series and locations of instances fallen into Nodes 5 and 7 of the tree depicted in Fig. 11

**Table 9** Ablation analysis: average running times (s) measured for SPALT and its variants without pruning (SPALT-NP), and without pruning and spatio-temporal locality (SPALT-NP-NS)

| | *UKPV dataset* | | | *SDWPF dataset* | | | *WPP dataset* | | |
|---|---|---|---|---|---|---|---|---|---|
| *Time (s)* | h=6 | h=12 | h=18 | h=6 | h=12 | h=18 | h=6 | h=12 | h=18 |
| SPALT | 169.40 | 170.66 | 173.58 | 168.72 | 209.50 | 242.25 | 176.72 | 210.98 | 249.49 |
| SPALT-NP | 110.45 | 118.59 | 133.66 | 142.44 | 176.50 | 210.63 | 156.99 | 188.95 | 227.47 |
| SPALT-NP-NS | 32.66 | 41.74 | 55.39 | 65.37 | 98.90 | 132.16 | 74.84 | 107.11 | 145.94 |

**Table 10** Average percentage of leaf nodes in which spatio-temporal locality features were introduced by SPALT

| | h = 6 | h = 12 | h = 18 |
|---|---|---|---|
| UKPV | 64.8% | 66.4% | 70.2% |
| SWDPF | 76.9% | 64.5% | 68.7% |
| WPP | 75.8% | 71.4% | 77.2% |

**Table 11** Analysis of the effect on the results (with $h = 12$) of the weighting strategy adopted on distances for the computation of spatio-temporal locality features in SPALT

| *UKPV dataset* | | | *SDWPF dataset* | | | *WPP dataset* | | |
|---|---|---|---|---|---|---|---|---|
| MAE | RMSE | RSE | MAE | RMSE | RSE | MAE | RMSE | RSE |
| *Linear distance weighting - Equations* (2) *and* (3) | | | | | | | | |
| 10.13 ± 4.9 | 20.83 ± 9.0 | 0.187 ± 0.09 | 125.5 ± 40.1 | 190.7 ± 51.0 | 0.344 ± 0.14 | 0.489 ± 0.11 | 0.850 ± 0.16 | 0.248 ± 0.04 |
| *Quadratic distance weighting - Equations* (2) *and* (10) | | | | | | | | |
| 10.12 ± 4.9 | 20.83 ± 8.9 | 0.188 ± 0.09 | 125.5 ± 40.0 | 190.7 ± 51.0 | 0.344 ± 0.14 | 0.489 ± 0.10 | 0.850 ± 0.16 | 0.248 ± 0.04 |

**Table 12** Analysis of the effect on the results (with $h = 12$) of the number of bins (*b*) used by the quantile binning strategy in SPALT

| *UKPV dataset* | | | *SDWPF dataset* | | | *WPP dataset* | | |
|---|---|---|---|---|---|---|---|---|
| MAE | RMSE | RSE | MAE | RMSE | RSE | MAE | RMSE | RSE |
| $b = 10$ | | | | | | | | |
| 10.06 ± 4.8 | 20.66 ± 8.7 | 0.184 ± 0.08 | 125.8 ± 40.4 | 191.0 ± 51.3 | 0.345 ± 0.14 | 0.49 ± 0.11 | 0.85 ± 0.16 | 0.250 ± 0.04 |
| $b = 25$ (*default*) | | | | | | | | |
| 10.13 ± 4.9 | 20.83 ± 9.0 | 0.187 ± 0.09 | 125.5 ± 40.1 | 190.7 ± 51.0 | 0.344 ± 0.14 | 0.49 ± 0.11 | 0.85 ± 0.16 | 0.248 ± 0.04 |
| $b = 50$ | | | | | | | | |
| 10.11 ± 4.9 | 20.80 ± 9.0 | 0.186 ± 0.08 | 125.4 ± 40.0 | 191.1 ± 51.1 | 0.346 ± 0.14 | 0.49 ± 0.10 | 0.85 ± 0.16 | 0.250 ± 0.04 |

$$C[\alpha, \beta] = 1 - \frac{D[\alpha, \beta]^2}{max(D^{\circ 2})} \quad (10)$$

where $D^{\circ 2}$ is the application of the Hadamard quadratic power operator to $D$, namely, $D^{\circ 2} = D \odot D$. This weighting function should in principle reduce more heavily the influence of very distant locations, with respect to the linear weighting. The obtained results are reported in Table 11.

Moreover, we also analyze the influence of the parameter $b$ that determines the number of bins of the quantile binning adopted by SPALT. Specifically, still focusing on the intermediate time horizon ($h = 12$), we ran additional experiments with $b = 10$ and $b = 50$, in comparison with the value adopted by default ($b = 25$). The results of such comparisons are reported in Table 12.

From the results, it is possible to observe that both the distance function and the value of $b$ do not significantly influence the results. This means that SPALT is quite robust to the value of such parameters, and that the proposed configuration represents a good trade-off between computational efforts and predictive performances.

Altogether, these results prove that SPALT as a whole provides significant advantages both in terms of predictive errors and in terms of model complexity, and that it can be considered as a state-of-the-art method for performing multi-step predictions exploiting the spatio-temporal locality of data generated from geodistributed sensors.

## 5 Conclusion

In this paper, we introduced SPALT, a novel multi-step forecasting method, based on linear model trees, designed to capture both linear and non-linear dynamics in multi-time series data generated by geo-distributed sensors. Specifically, we extended linear model trees to account for spatio-temporal locality by leveraging the spatial proximity of sensors exhibiting similar trends. This was achieved by selectively enriching the linear models at the leaf nodes with additional features derived from historical trends observed at other locations within the same leaf node. Furthermore, we proposed a new pruning strategy based on Reduced Error Pruning (REP) that incorporates spatio-temporal locality into the tree simplification process.

Our experiments, conducted on three real-world datasets related to renewable energy production from renewable power plants, demonstrated the effectiveness of SPALT compared to various tree-based models and state-of-the-art neural network architectures capable of modeling both temporal and spatial dimensions. Moreover, our ablation study showed a statistically significant advantage of considering spatio-temporal locality aspects, as well as a significant reduction of the model complexity achieved by our pruning strategy, without affecting the predictive accuracy.

For future work, we will evaluate the effectiveness of the proposed method in other application domains other than the energy domain investigated in this paper. Additionally, we will explore its extension to ensemble methods, such as forests of linear model trees and extremely randomized linear model trees.

**Acknowledgements** This work was partially supported by the project FAIR - Future AI Research (PE00000013), Spoke 6 - Symbiotic AI, under the NRRP MUR program funded by the NextGenerationEU. It is also partially supported by the Italian MUR, PRIN 2022 Project “COCOWEARS” (Grant n. 2022T2XNJE - CUP: H53D23003650001) funded by the European Union - NextGenerationEU, under the National Recovery and Resilience Plan (NRRP) Mission 4 Component 2 Investment Line 1.1. The research of Annunziata D’Aversa is funded by a PhD fellowship within the framework of the Italian "POR Puglia FSE 2014-2020" – Axis X - Action 10.4 "Interventions to promote research and for university education" - PhD Project n. 1004.121 (CUP n. H99J21006620008).

**Author contributions** AD, GP and MC conceived the task and designed the solution from a methodological point of view. AD implemented the algorithms, ran the experiments and collected the results. GP, AD and MC interpreted and discussed the results. GP and MC supervised the whole research. All the authors contributed to the manuscript drafting and approved the final version of the manuscript.

**Funding** Open access funding provided by Università degli Studi di Bari Aldo Moro within the CRUI-CARE Agreement.

**Data availability** Two out of three datasets used in the experiments are publicly available at the following links: SDWPF dataset: https://aistudio.baidu.com/competition/detail/152; UKPV dataset: https://huggingface.co/datasets/openclimatefix/uk_pv.

## Declarations

**Competing interests** The authors declare no competing interests.

## Authors and Affiliations

**Annunziata D'Aversa[1] · Gianvito Pio[1,2] · Michelangelo Ceci[1,2,3]**

✉ Gianvito Pio
gianvito.pio@uniba.it

Annunziata D'Aversa
annunziata.daversa@uniba.it

Michelangelo Ceci
michelangelo.ceci@uniba.it

1 Dept. of Computer Science, University of Bari Aldo Moro, Via E. Orabona, 4, 70125 Bari, Italy

2 Data Science Lab, National Interuniversity Consortium for Informatics (CINI), Via Volturno, 58, 00185 Roma, Italy

3 Dept. of Knowledge Technologies, Jozef Stefan Institute, Jamova Cesta, 00185 Ljubljana, Slovenia